\documentclass[final]{cvpr}

\usepackage{times}
\usepackage{epsfig}
\usepackage{graphicx}
\usepackage{amsmath}
\usepackage{amssymb}
\usepackage{flushend}
\usepackage[tight,footnotesize]{subfigure}
\usepackage{enumitem}
\usepackage[breaklinks=true,colorlinks]{hyperref}

\def\cvprPaperID{4459} % *** Enter the CVPR Paper ID here
\def\confYear{CVPR 2021}

\begin{document}

%%%%%%%%% TITLE
\title{Rank-One Prior:  Toward Real-Time Scene Recovery}

\author{Jun Liu$^{1,2}$, Ryan Wen Liu$^3$, Jianing Sun$^{1,4}$\thanks{Corresponding author.}, Tieyong Zeng$^5$\\
$^1$School of Mathematics and Statistics, Northeast Normal University, Changchun, China\\
$^2$Key Laboratory of Applied Statistics of MOE at Northeast Normal University\\
$^3$School of Navigation, Wuhan University of Technology, Wuhan, China\\
$^4$Jilin National Applied Mathematics Center at Northeast Normal University\\
$^5$Department of Mathematics, The Chinese University of Hong Kong, Shatin, NT, Hong Kong\\
\centerline{\texttt{\large{\{liuj292,sunjn118\}@nenu.edu.cn, wenliu@whut.edu.cn, zeng@math.cuhk.edu.hk}}
}%Institution2\\
}

\pagestyle{empty}  % no page number for the second and the later pages
\maketitle
\thispagestyle{empty}

%%%%%%%%% ABSTRACT
\begin{abstract}
Scene recovery is a fundamental imaging task for several practical applications, e.g., video surveillance and autonomous vehicles, etc. To improve visual quality under different weather/imaging conditions, we propose a real-time light correction method to recover the degraded scenes in the cases of sandstorms, underwater, and haze. The heart of our work is that we propose an intensity projection strategy to estimate the transmission. This strategy is motivated by a straightforward rank-one transmission prior. The complexity of transmission estimation is $O(N)$ where $N$ is the size of the single image. Then we can recover the scene in real-time. Comprehensive experiments on different types of weather/imaging conditions illustrate that our method outperforms competitively several state-of-the-art imaging methods in terms of efficiency and robustness.
\end{abstract}
%
%%%%%%%%% BODY TEXT
%
%
\section{Introduction}
\begin{figure}[!t]
\begin{minipage}[b]{.32\linewidth}
\centering
\centerline{\includegraphics[width=2.65cm]{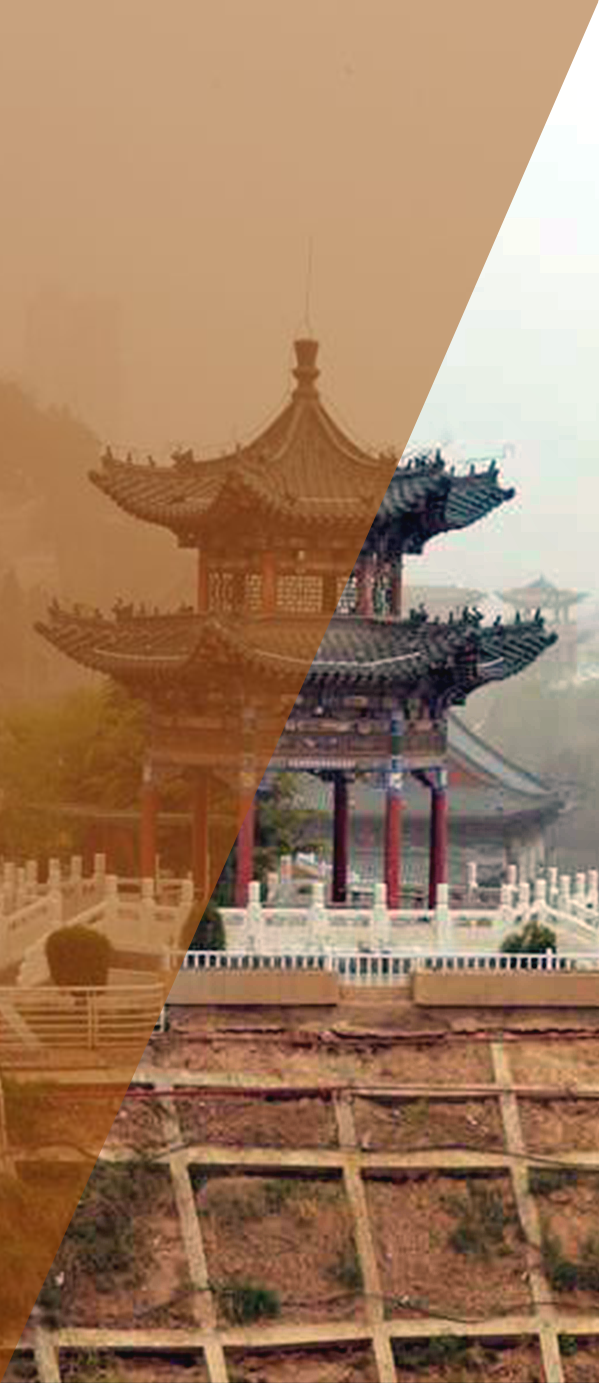}}
\centerline{\footnotesize{(a) sandstorm image}  }\medskip
\end{minipage}
\begin{minipage}[b]{.33\linewidth}
\centering
\centerline{\includegraphics[width=2.65cm]{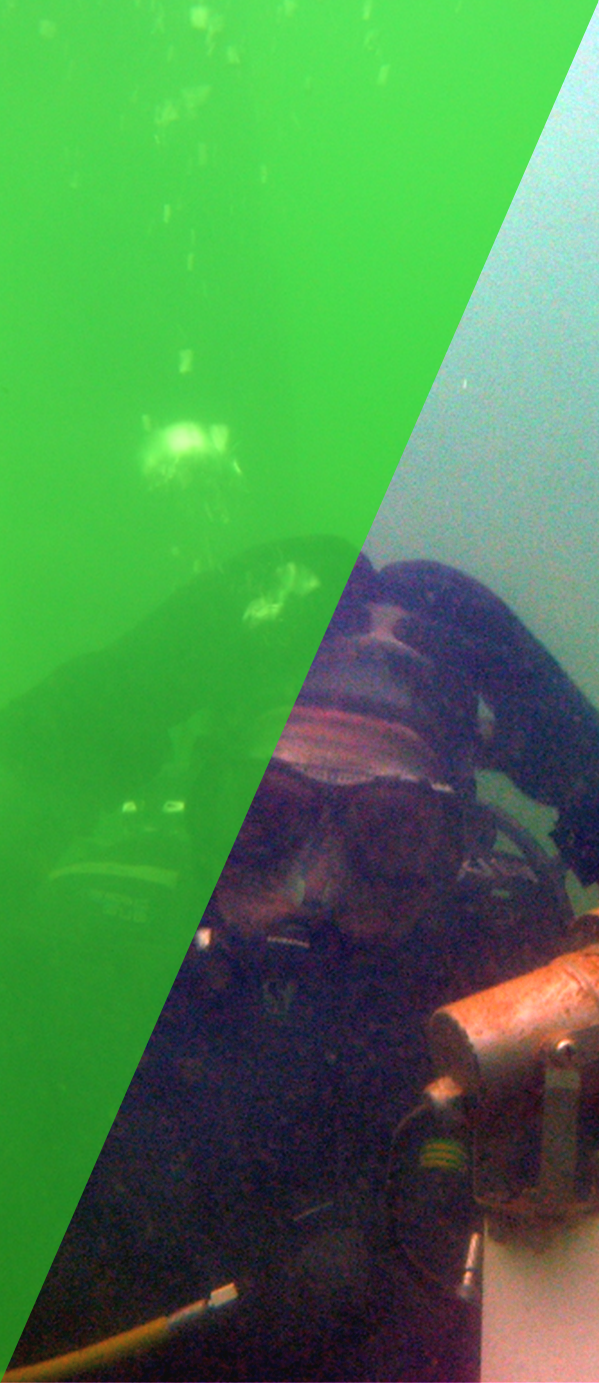}}
\centerline{\footnotesize{(b) underwater image}  }\medskip
\end{minipage}
\begin{minipage}[b]{.33\linewidth}
\centering
\centerline{\includegraphics[width=2.65cm]{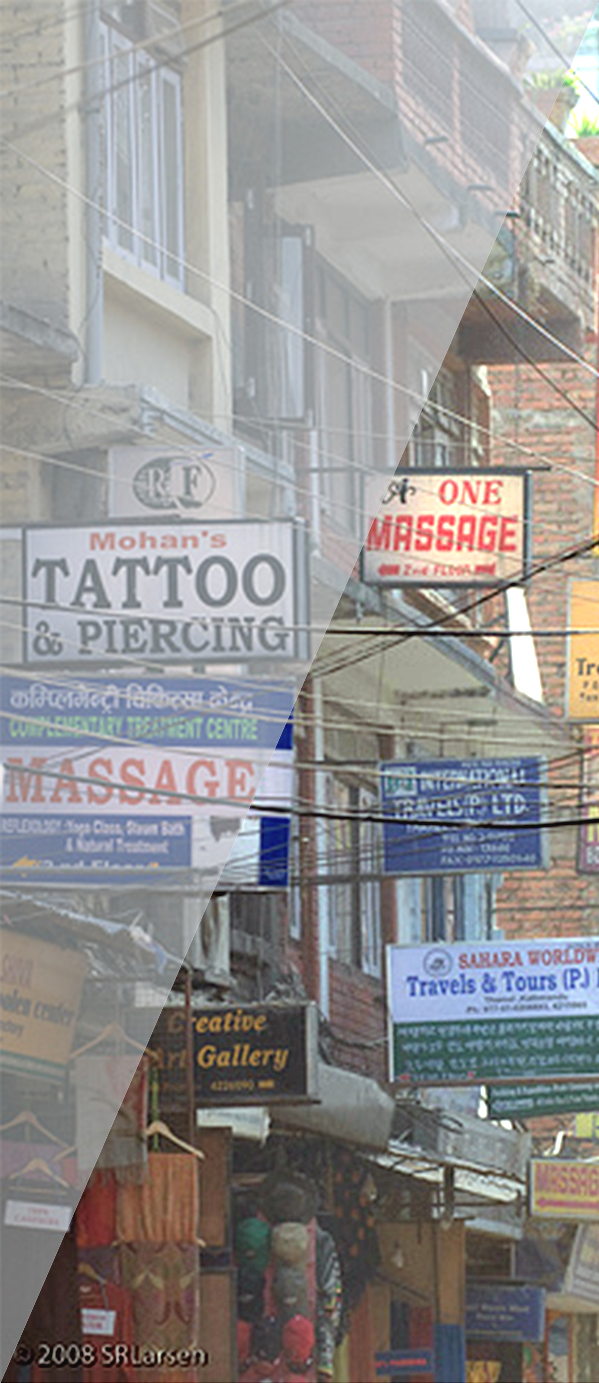}}
\centerline{\footnotesize{(c) hazy image} }\medskip
\end{minipage}
\caption{Example of different imaging conditions. The upper triangles in (a)-(c) are degraded patterns and the corresponding patterns in the lower triangles are restored by our method.}
\label{fig:degradation_types}
\end{figure}

An image is worth one thousand words. However, an image or a video captured in a turbid medium would not provide sufficient and correct visual information. Figure \ref{fig:degradation_types} shows three images captured in  turbid media, such as sandstorms, underwater, and haze. Such kind of images suffer from severe contrast and color alteration or degradation. These changes increase the difficulty in many computer vision tasks, including object tracking, marine celestial navigation, pattern recognition, and semantic segmentation. Hence, recovering the correct scene from the degraded observation is an essential and fundamental task in computer vision.  

For sandstorm images, underwater images, and hazy images, the degradation is generally due to light absorption and scattering \cite{he2010single,sulami2014automatic,fang2014single,fu2014fusion,li2020underwater}. More specifically, 1) due to the physical properties of the environmental medium, it will scatter the visible light formed by the relaxation of multiple spectra such that the scattered light source and the initial visible light source together constitute the environmental light source; 2) the environmental medium also absorbs the visible light, and the imaging process is completed by the imaging equipment to obtain the emitted light irradiation of the imaging scene. The acquired radiation intensity will then be attenuated under the environmental mechanism, and the attenuation degree depends on the distance between the imaging scene and the imaging equipment.  Usually, there are conventional names for image restorations, such as sand-dust/underwater image enhancement and image dehazing. Although the visual degradation appears differently, they share similar features: color distortion, low contrast, and low visibility. 

The physical model that is widely used to describe the formation of an image suffered from light transmission hazard \cite{schechner2007regularized,peng2018generalization,gao2019underwater} is often defined as follows:
\begin{equation}\label{eq:IMF}
    \mathbf{I}(x) =  t(x)  \mathbf{J}(x) + \tilde{t}(x) \mathbf{A}, 
\end{equation}
where $\mathbf{I}$ is the observed degraded image, $\mathbf{J}$ is the scene radiance, and $\mathbf{A}$ is the global ambient light. Moreover,  $\tilde{t}(x) + t(x) = 1$,  $t$ is the medium transmission describing the portion of the light reaching the camera, and $\tilde{t}(x)$ is the transmission describing the portion of the ambient scattered light that influences imaging. In 1924, this model was first proposed by Koschmieder \cite{koschmieder1924theorie}. To recover the scene radiance $\mathbf{J}$, we need both the transmission $t$ (or $\tilde{t})$ and the ambient light $\mathbf{A}$. This task is challenging since both variables are unknown. Furthermore, this problem is under-determined since we usually have only one single degraded image. 

When the environmental transmitter is homogeneous, the $t(x)$ in Eq.  (\ref{eq:IMF}) can be expressed as follows:
\begin{equation}\label{rad_inten}
	t(x) = e^{-\beta d(x)},
\end{equation}
where $\beta$ is the scattering coefficient of the ambient medium and $d(x)$ is the scene depth.  

Nevertheless, a single image scene recovery has received much attention and obtained great progress and new progress. Narasimhan and Nayar \cite{narasimhan2002vision} derived geometric constraints on scene color changes and then developed algorithms to recover scene colors as they would appear on a clear day. In \cite{narasimhan2003contrast}, the same authors further presented a physics-based model that described the appearances of scenes in uniform bad weather conditions and they proposed a fast algorithm to restore scene contrast. Fu et al. \cite{fu2014fusion} proposed a sandstorm image enhancement approach based on fusion principles. Wang et al. \cite{wang2020fast} proposed a fast color balance and multi-path fusion method for sandstorm image enhancement. Based on the assumption that the transmission and surface shading are locally uncorrelated, Fattal \cite{fattal2008single} estimated the surface albedo and then inferred the medium transmission. With two observations, i.e., haze-free images have more contrast than images plagued by bad weather; and airlight tends to be smooth, Tan \cite{tan2008visibility} proposed to remove the haze by maximizing the local contrast of the restored image. Fang et al. \cite{fang2020variational} proposed to handle the hazy image in YUV color space with two priors and they showed that most of the chrominance information of the image can be well preserved. Generally, these preset conditions can be viewed as priors of the latent clean images.  Referring to priors, the assumption of dark channel prior (DCP), proposed by He et al. \cite{he2010single}, could be regarded as the best-known one. The DCP is based on the statistics of haze-free images. Afterwards, numerous DCP-based methods \cite{meng2013efficient,huang2014visibility,chiang2011underwater,shu2019variational,tang2014investigating,drews2016underwater} have been proposed for image restoration, such as sandstorm, underwater, and hazy images. In \cite{golts2019unsupervised}, Golts et al. proposed a dehazing method of unsupervised training of deep neural networks based on DCP loss. Note that the DCP has also been successfully applied in blur kernel estimation \cite{pan2016blind}. However, the DCP-based methods may fail if the haze-free images do not contain zero-intensity pixels. Unlike the DCP that assumes zero minimal value in local patches, Fattal \cite{fattal2014dehazing} proposed a single image dehazing method based on the color-lines pixel regularity in natural images. There are also some other effective priors, to name a few, color attenuation prior \cite{zhu2015fast}, non-local prior \cite{berman2016non}, color ellipsoid prior \cite{bui2017single}, gradient channel prior \cite{kaur2020color}, and gamma correction prior \cite{ju2019idgcp}, etc.

In recent years, the convolutional neural network (CNN) based methods for single image restoration have become popular.  Cai et al. \cite{cai2016dehazenet} proposed a deep neural network (DehazeNet) for transmission map and then used the conventional method to estimate atmospheric light. A multi-scale version of Dehazenet \cite{ren2016single} was trained to estimate the transmission map. In \cite{li2017aod}, Li et al. proposed to directly restore the latent sharp image from a hazy image through a light-weight CNN (AOD-Net). Yang and Sun \cite{yang2018proximal} proposed a proximal Dehaze-Net by incorporating the haze imaging model, dark channel, and transmission priors into a deep architecture. Jamadandi and Mudenagudi \cite{jamadandi2019exemplar} proposed a deep learning framework to enhance the underwater image with wavelet corrected transformations. Li et al. \cite{li2020underwater} adopted the CNN network to perform underwater image enhancement based on the assumption of underwater scene prior. Li et al. \cite{li2019underwater} constructed an underwater image enhancement benchmark and proposed an underwater image enhancement network (Water-Net) trained on their proposed model. However, these supervised learning-based imaging results essentially depend on the diversity and volume of collected datasets. The generative adversarial network (GAN), an unsupervised learning method, has achieved significant success toward scene recovery \cite{deng2020hardgan,liu2019mlfcgan,qu2019enhanced}. It is capable of generating realistic-looking synthetic images and enhance the visual quality under different weather/imaging conditions.

We can briefly divide the aforementioned methods into two main categories: traditional physical property-based methods (such as preset priors) and data-driven based methods (end-to-end network training methods). In this paper, to make scene recovery easier and more flexible, we focus on the first case due to its simplicity, stability, and flexibility. The contributions of this work are as follows:

\begin{itemize} 
	\item 
	We propose a new and straightforward method for estimating the transmission based on the rank-one prior. This effective method achieves state-of-the-art performance for various applications under different weather/imaging conditions.
	\item 
	The complexity of transmission estimation is $O(N)$ where $N$ is the size of the single image. The proposed method is then super fast and the GPU acceleration achieves $4\sim8$ times faster than the CPU computing. 
	\item 
	The implementation of our method is very easy since there is no iteration step. Furthermore, we expect that our method can be applied in other related tasks.
\end{itemize}

\section{Rank-One Transmission Prior}
Rank-one transmission prior is based on the following observation for outdoor degradation images:  in  most of the regions except the light source area, the imaging scene is covered by spatially homogenous light. The thickness of scattered light depends on the scenery depth, or the light transmission. The transmission $\tilde{t}$ has strong correlation with scattered light. If the spectrum of scattered light is given, the transmission can be characterized by their correlation coefficients. For further discussion, let us describe their correlation to be linear.  Mathematically, if we stack the transmission $\tilde{t}(x) \in \mathbb{R}^{1\times 3}$ into a matrix $\tilde{\mathbf{T}}\in \mathbb{R}^{r\times 3}$ ($r = mn$, if the image $\mathbf{I}\in \mathbb{R}^{m\times n \times 3}$), the matrix $\tilde{\mathbf{T}}$ should be a rank-one matrix. Specifically, if $\mathbf{J}$ is an outdoor scene image except for the ambient light source, the transmission $\tilde{t}(x)$ can be represented as follows:
\begin{equation}\label{rank1formula}
	\tilde{\mathbf{T}} = \mathbf{C} \mathbf{I}_u, 
\end{equation}
where $\mathbf{C}\in\mathbb{R}^{r}$ is a coefficient vector and $\mathbf{I}_u\in\mathbb{R}^{1\times 3}$ is the transmission basis of $\tilde{t}$. We call this observation rank-one transmission prior.

\subsection{Rank-one prior validation}
In this subsection, we provide statistical support for the correctness of our rank-one prior. We carry out a statistical experiment on several novel datasets named I-Haze \cite{Ancuti2018ihaze}, O-Haze \cite{Ancuti2018ohaze}, and Dense-Haze \cite{Ancuti2019Dense}, respectively. These datasets are of real hazy images obtained in indoor and outdoor environments with ground truth. I-Haze and O-Haze have been employed in the dehazing challenge of the NTIRE 2018 CVPR workshop \cite{ancuti2018ntire}. Dense-Haze has been used in the dehazing challenge of the NTIRE 2019 CVPR workshop \cite{Ancuti19ntire}. Unlike other dehazing databases, hazy images in I-Haze, O-Haze, Dense-Haze are generated using real haze produced by a professional haze machine. We collect 30 pairs of real hazy and corresponding haze-free images of various indoor scenes from I-Haze, 45 pairs of different outdoor scenes from O-Haze, and 55 pairs of dense homogeneous hazy images and haze-free images from Dense-Haze.

To give a more comprehensive demonstration, we compute the transmission map $t \in \mathbb{R}^{m\times n\times 3}$ by two different ways, and stack the third-order slice of  $t$ into a matrix $\mathbf{T} \in \mathbb{R}^{mn \times 3}$. We compute the singular value decomposition (SVD) of $\mathbf{T}$ and verify that the best rank-1 approximation $\mathbf{T}_1$ takes the major energy of $\mathbf{T}$, i.e., the energy percentage \cite{ji2016rank} $r$ of the obtained rank-one matrix $\mathbf{T}_1$ satisfies $r = \frac{\|\mathbf{T}_1\|_F^2}{\|\mathbf{T}\|_F^2} = \frac{\sigma_1^2}{\sum_i^3 \sigma^2_i}\geq  90\%$ with $\sigma_i, i=1,2,3$ the singular value of $\mathbf{T}$.

\begin{itemize}
	\item {\bf Case 1}. Using the physical image formation model. Figure \ref{Fig:case1} (a) shows the energy percentage of the obtained rank-one approximations among 130 different hazy images. From this real case, we can observe that for over $96\%$ images,  the rank-one transmission energy is higher than $90\%$ of the total energy, which ensures the correctness of the proposed prior.
	\begin{figure}
		\centering
		\subfigure[]{
		\includegraphics[scale=0.23]{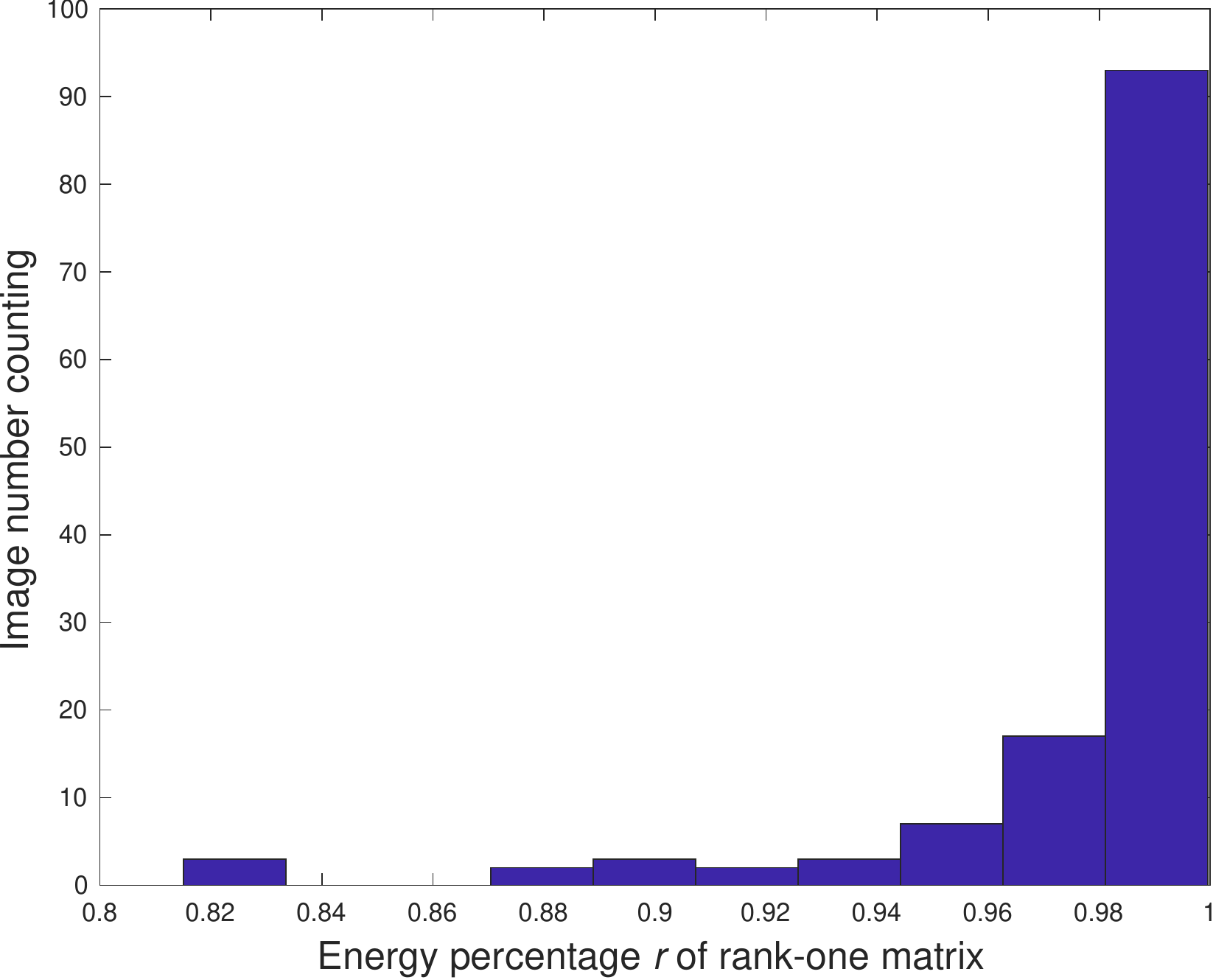} }
		\subfigure[]{
		\includegraphics[scale=0.23]{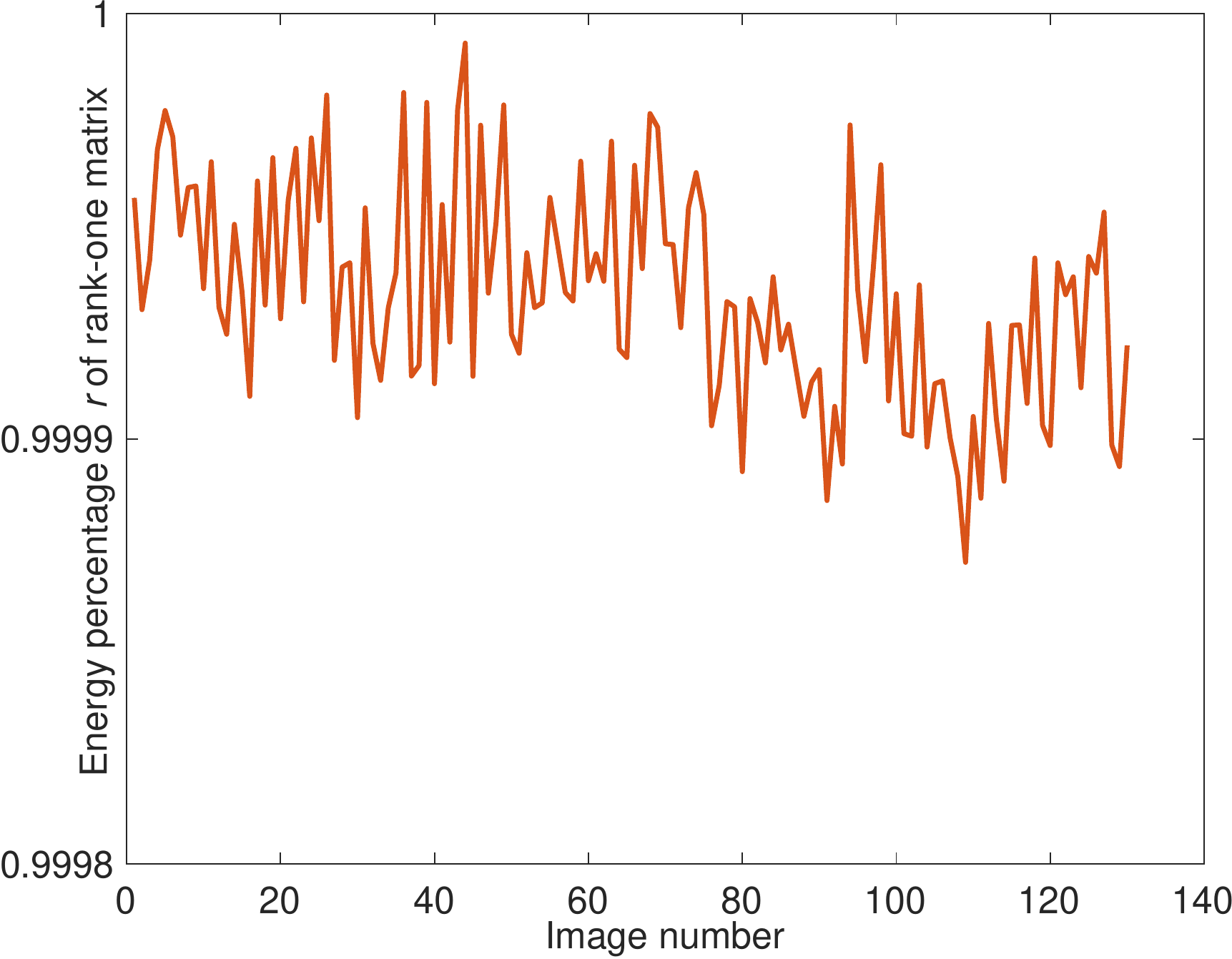} }
		\caption{The energy percentage of the obtained rank-one approximations. (a) The transmission map is computed by using the physical image formation model. (b) The transmission map is computed by using an existing transmission estimation method \cite{shu2019variational}.}
		\label{Fig:case1}
	\end{figure}
	\item {\bf Case 2}. Using an existing transmission estimation method \cite{shu2019variational}. In this case, we first compute the transmission map by Shu et al.'s method for each hazy image, and then stack the third-order slice of  $t$ into a matrix $\mathbf{T} \in \mathbb{R}^{mn\times 3}$. Similar to Case 1, we display the energy ratio of rank-one approximation in Figure \ref{Fig:case1} (b). All rank-one approximations' energy percentages achieve more than $99.98\%$. This result further proves that our proposed rank-one prior is valid. 
 \end{itemize}

\begin{figure*}
	\begin{center}
	\includegraphics[width=1\linewidth]{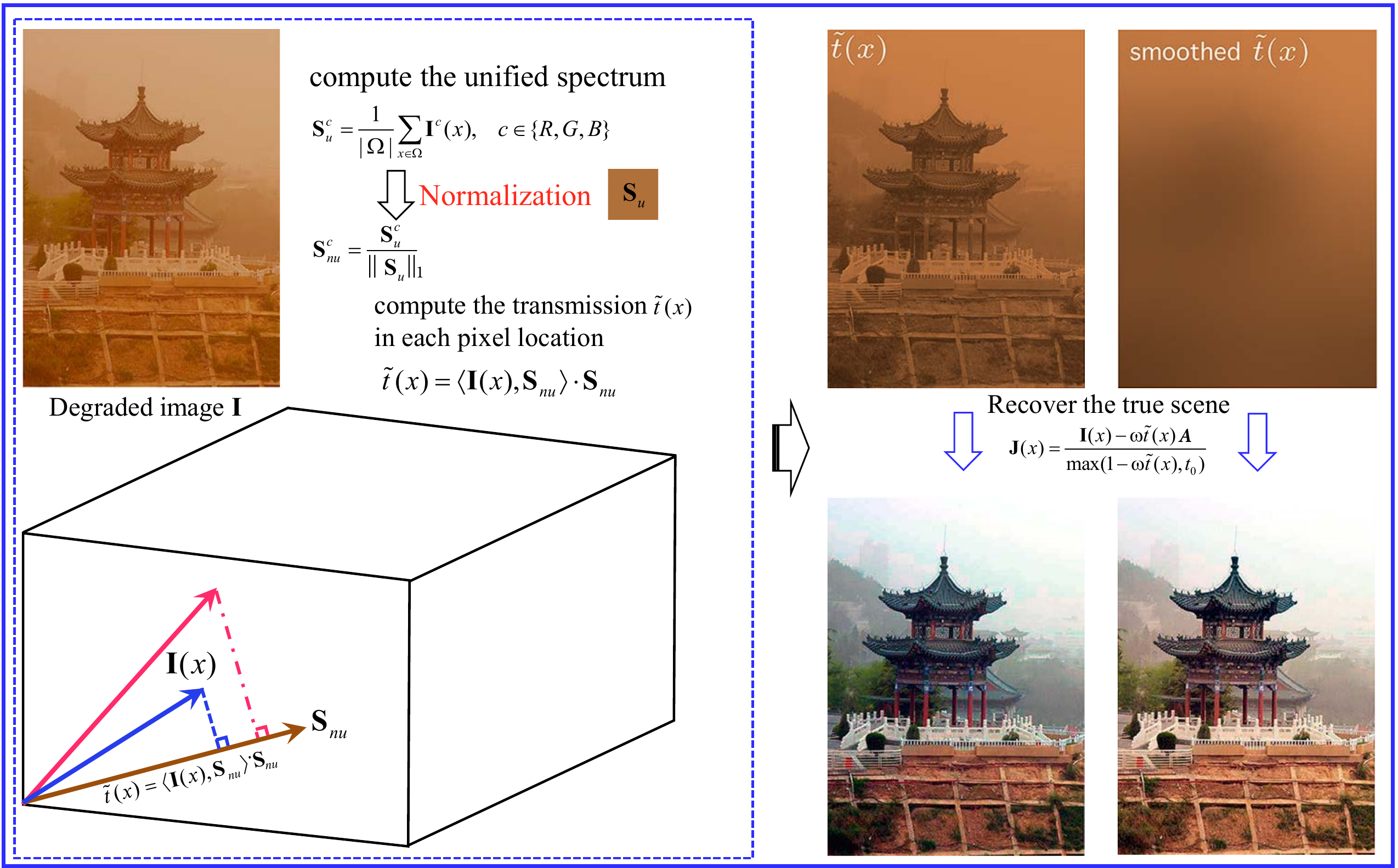}
	\end{center}
 	\caption{Flowchart of the proposed method. $\mathbf{S}_u$ denotes the unified radiance and $\mathbf{S}_{nu}$ denotes the unified spectrum. The projection of $\mathbf{I}(x)$ onto the unified spectrum is the scatter light transmission $\tilde{t}(x)$. 
	  }
	\label{fig:flowchart}
\end{figure*}

 \begin{figure*}[t]
	\footnotesize
	\begin{minipage}[b]{.12\linewidth}
		\centering
		\centerline{\includegraphics[width=2.15cm]{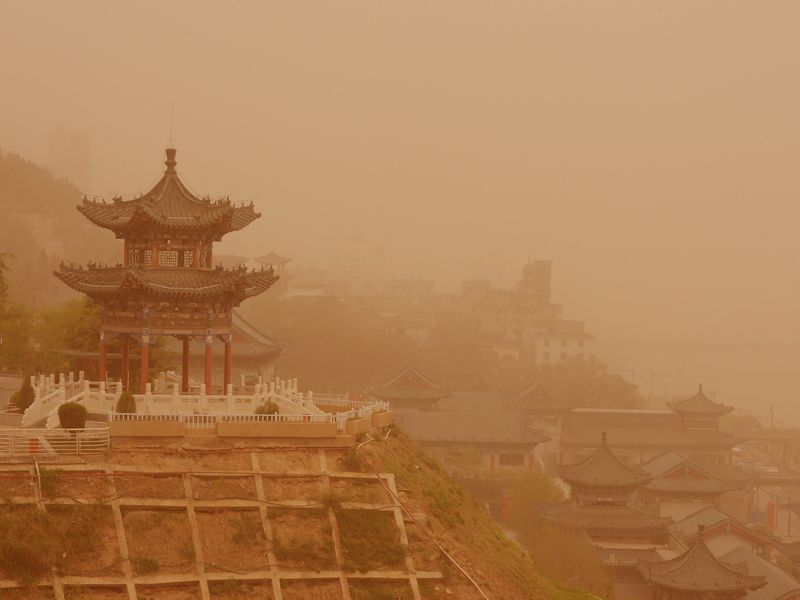}}
\vspace{0.01cm}
	\end{minipage}
	\begin{minipage}[b]{.12\linewidth}
		\centering
		\centerline{\includegraphics[width=2.15cm]{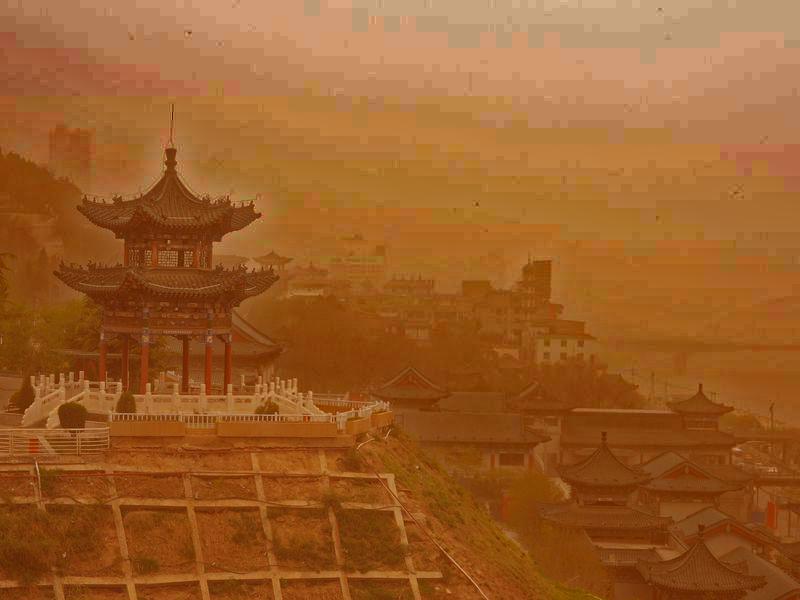}}
\vspace{0.01cm}
	\end{minipage}
	\begin{minipage}[b]{0.12\linewidth}
		\centering
		\centerline{\includegraphics[width=2.15cm]{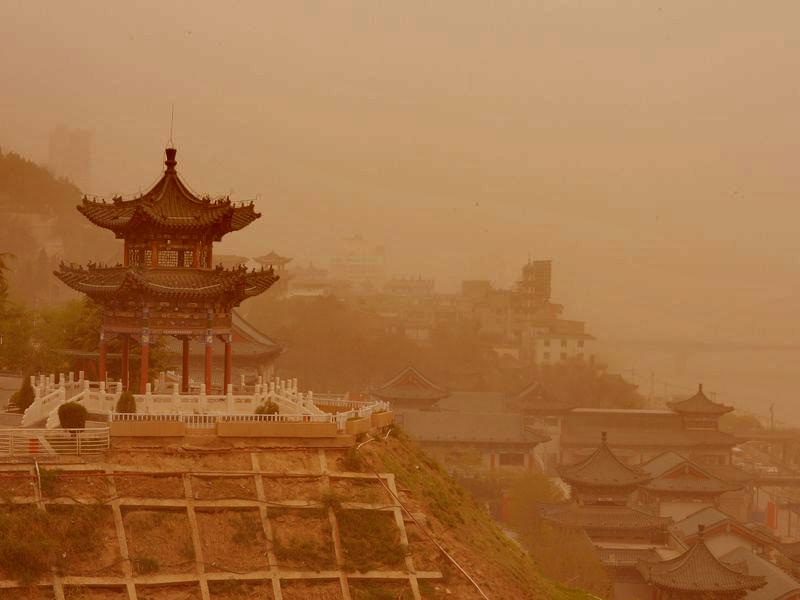}}
\vspace{0.01cm}
	\end{minipage}
	\begin{minipage}[b]{0.12\linewidth}
		\centering
		\centerline{\includegraphics[width=2.15cm]{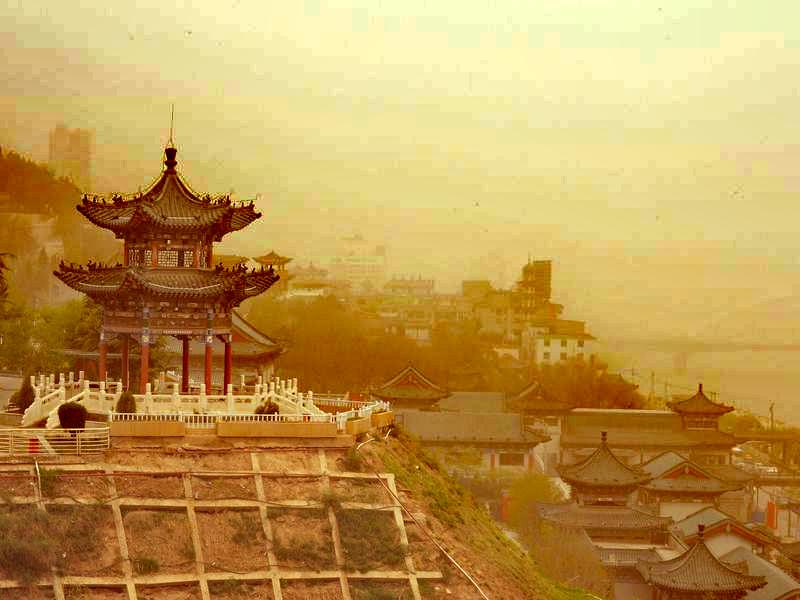}}
\vspace{0.01cm}
	\end{minipage}
	\begin{minipage}[b]{0.12\linewidth}
		\centering
		\centerline{\includegraphics[width=2.15cm]{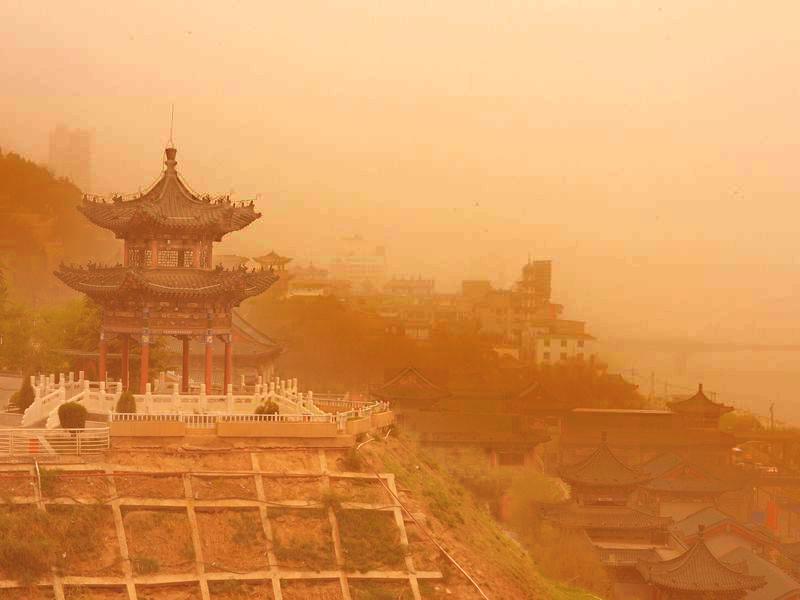}}
\vspace{0.01cm}
	\end{minipage}
	\begin{minipage}[b]{0.12\linewidth}
	\centering
	\centerline{\includegraphics[width=2.15cm]{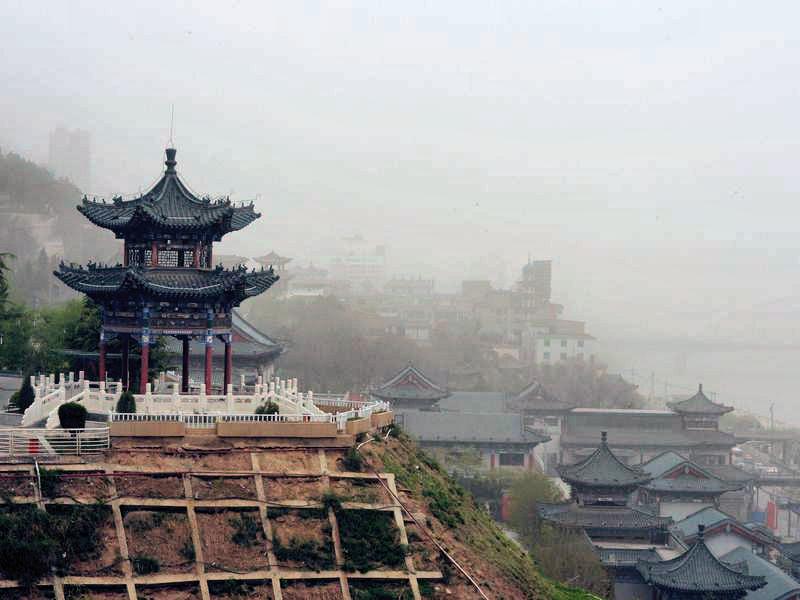}}
	\vspace{0.01cm}
\end{minipage}
\begin{minipage}[b]{.12\linewidth}
	\centering
	\centerline{\includegraphics[width=2.15cm]{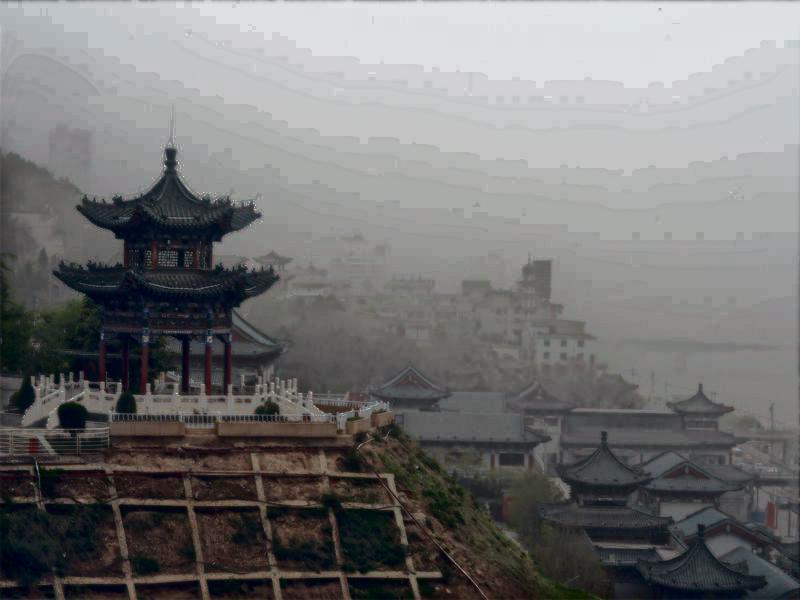}}
	\vspace{0.01cm}
\end{minipage}
    \begin{minipage}[b]{0.12\linewidth}
    	\centering
    	\centerline{\includegraphics[width=2.15cm]{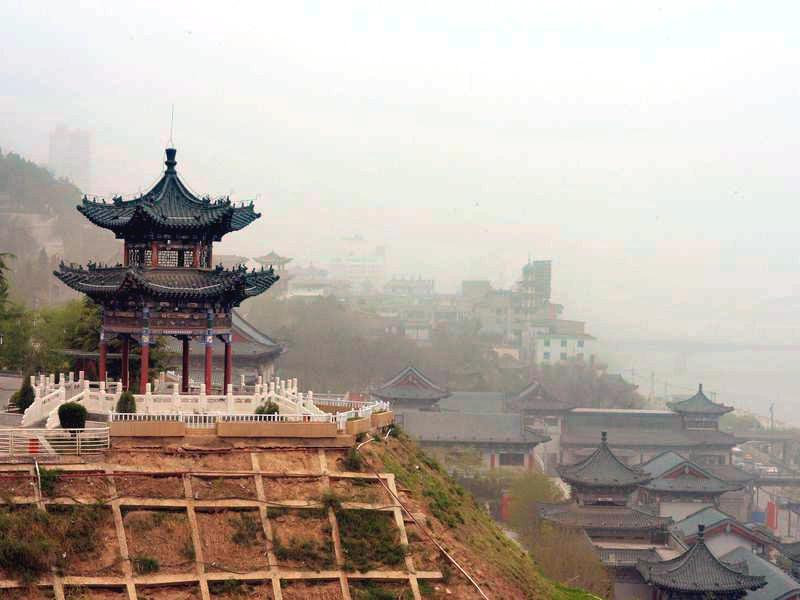}}
    	\vspace{0.01cm}
    \end{minipage}
	\\
	\begin{minipage}[b]{.12\linewidth}
	\centering
	\centerline{\includegraphics[width=2.15cm]{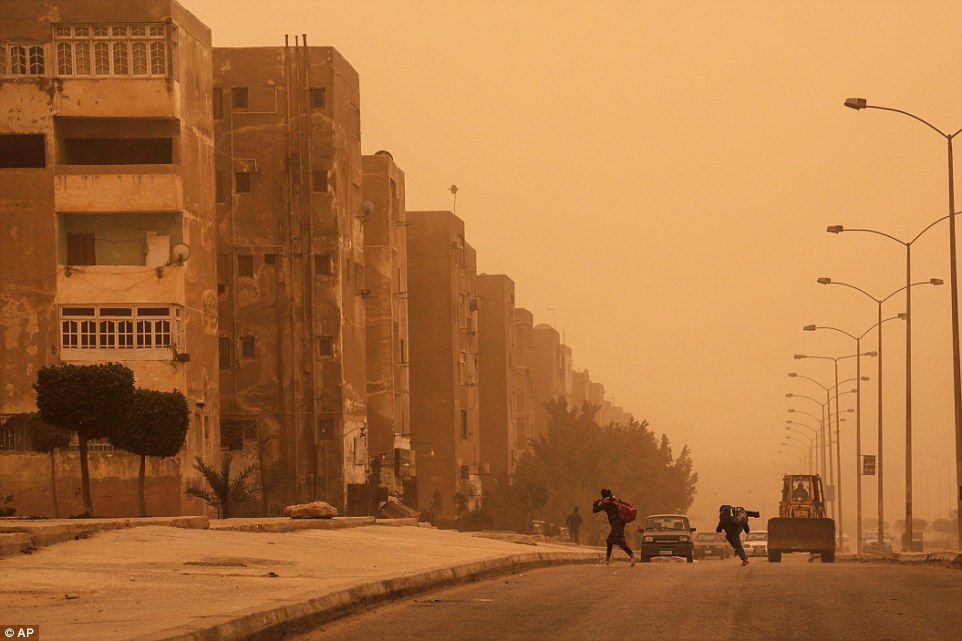}}
	\vspace{0.01cm}
\end{minipage}
\begin{minipage}[b]{.12\linewidth}
	\centering
	\centerline{\includegraphics[width=2.15cm]{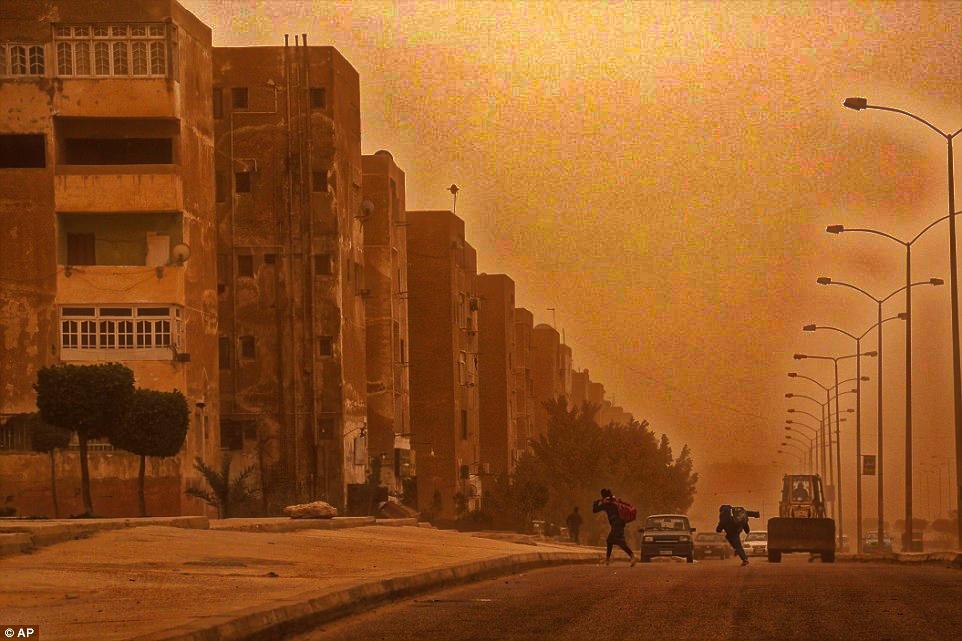}}
	\vspace{0.01cm}
\end{minipage}
\begin{minipage}[b]{0.12\linewidth}
	\centering
	\centerline{\includegraphics[width=2.15cm]{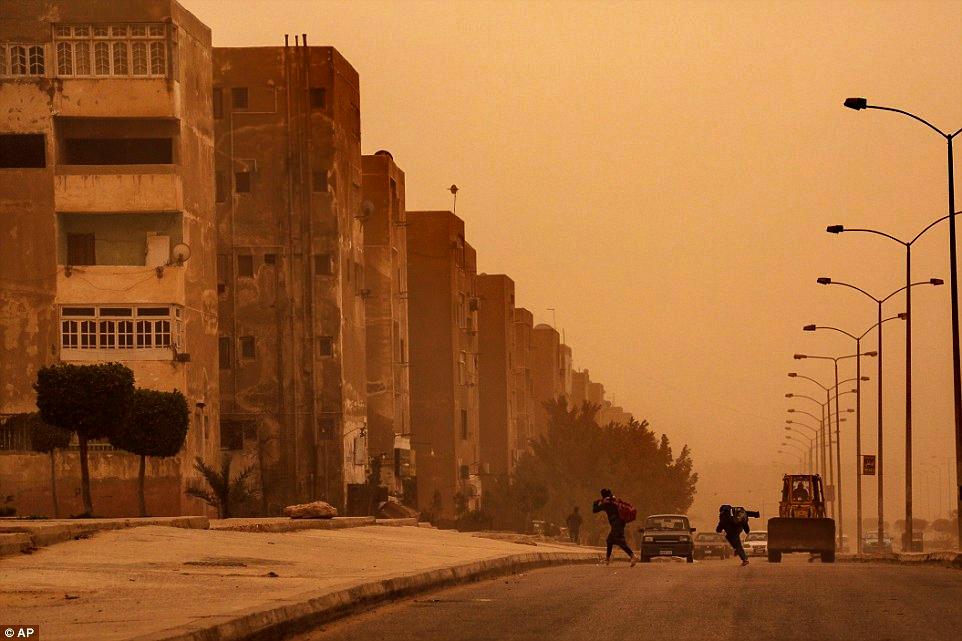}}
	\vspace{0.01cm}
\end{minipage}
\begin{minipage}[b]{0.12\linewidth}
	\centering
	\centerline{\includegraphics[width=2.15cm]{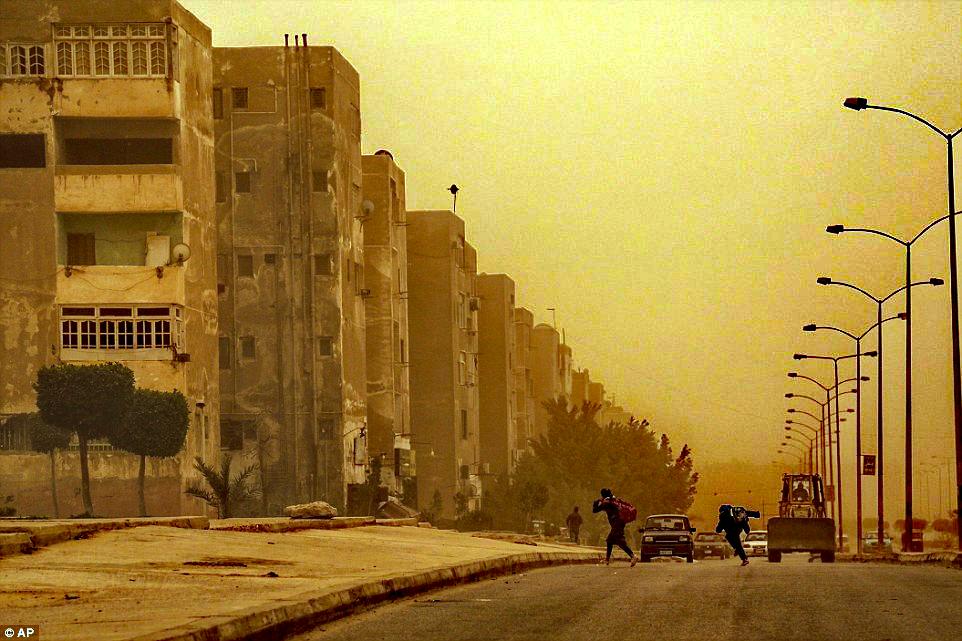}}
	\vspace{0.01cm}
\end{minipage}
\begin{minipage}[b]{0.12\linewidth}
	\centering
	\centerline{\includegraphics[width=2.15cm]{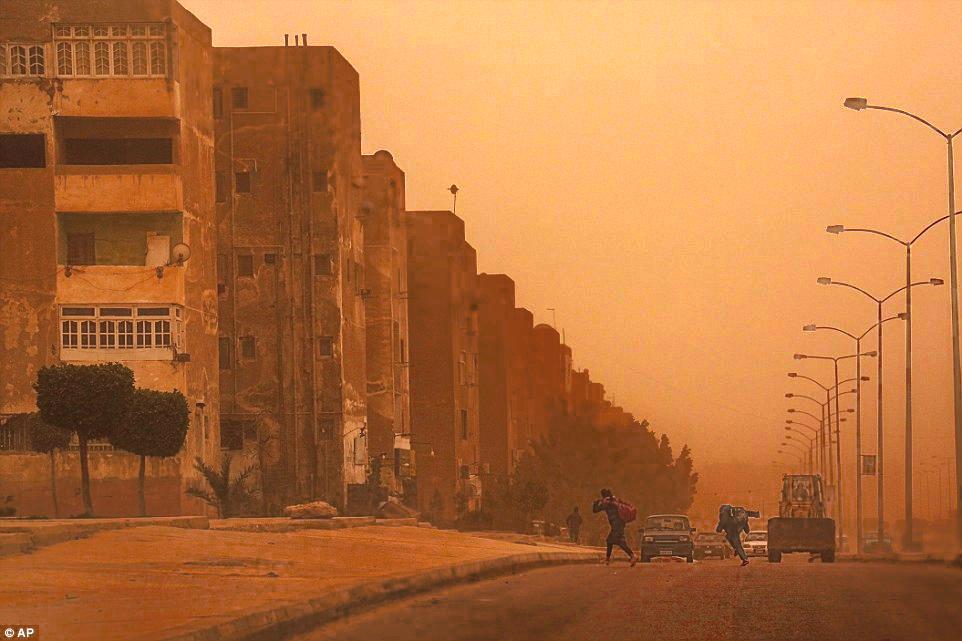}}
	\vspace{0.01cm}
\end{minipage}
\begin{minipage}[b]{0.12\linewidth}
	\centering
	\centerline{\includegraphics[width=2.15cm]{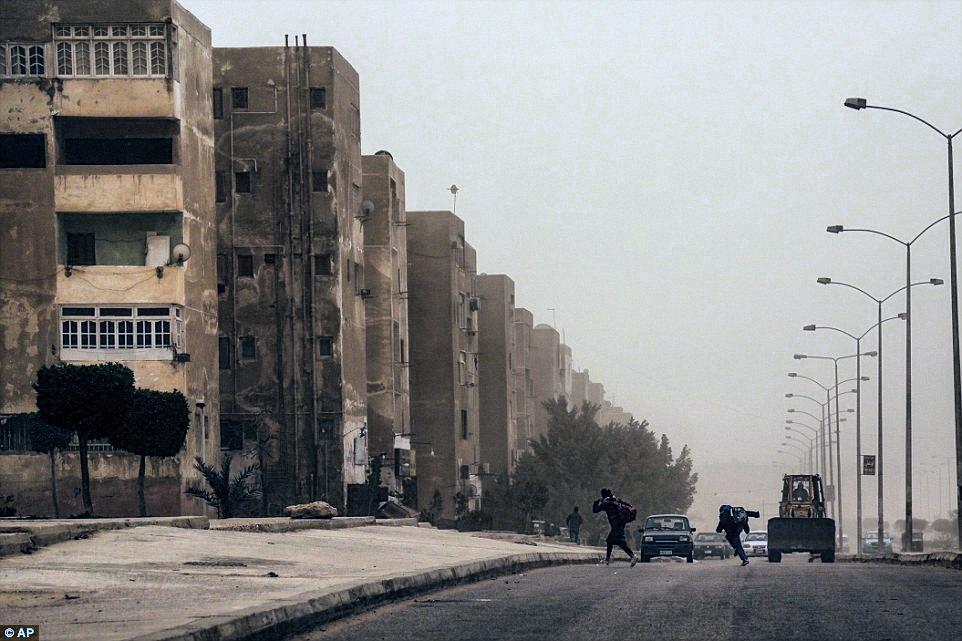}}
	\vspace{0.01cm}
\end{minipage}
\begin{minipage}[b]{.12\linewidth}
	\centering
	\centerline{\includegraphics[width=2.15cm]{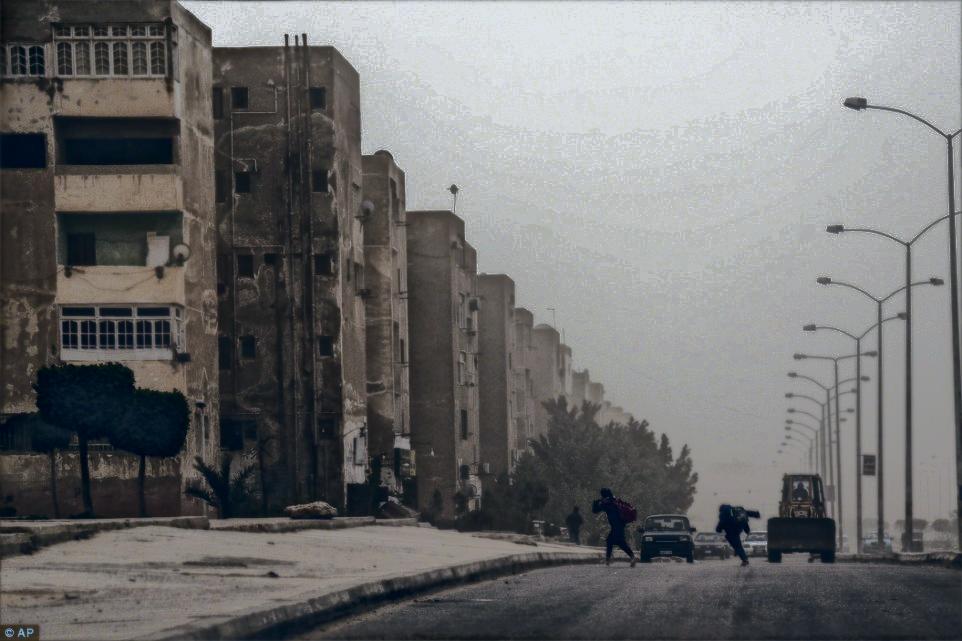}}
	\vspace{0.01cm}
\end{minipage}
\begin{minipage}[b]{0.12\linewidth}
	\centering
	\centerline{\includegraphics[width=2.15cm]{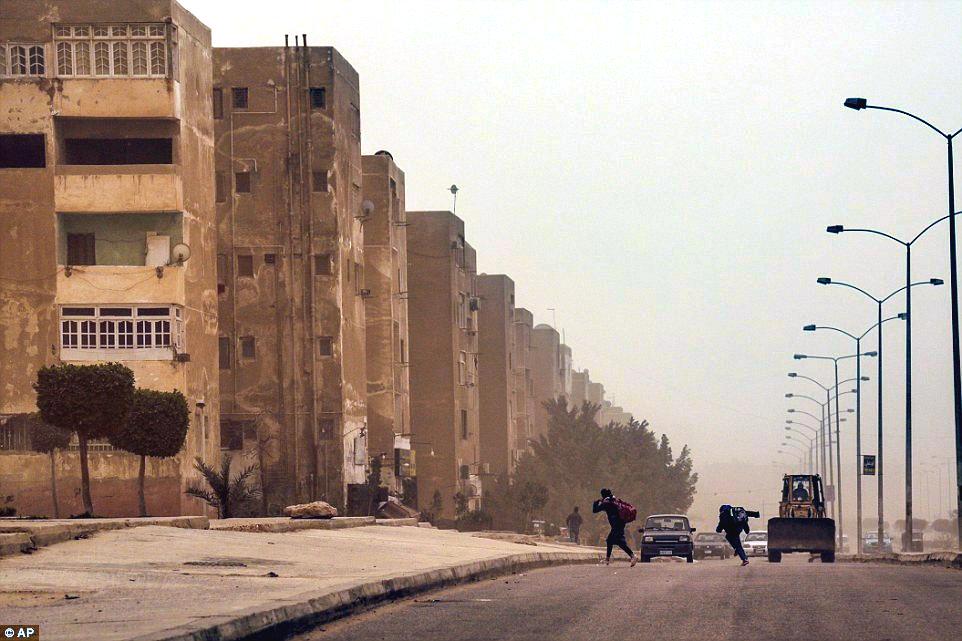}}
	\vspace{0.01cm}
\end{minipage}
	\\
\begin{minipage}[b]{.12\linewidth}
	\centering
	\centerline{\includegraphics[width=2.15cm]{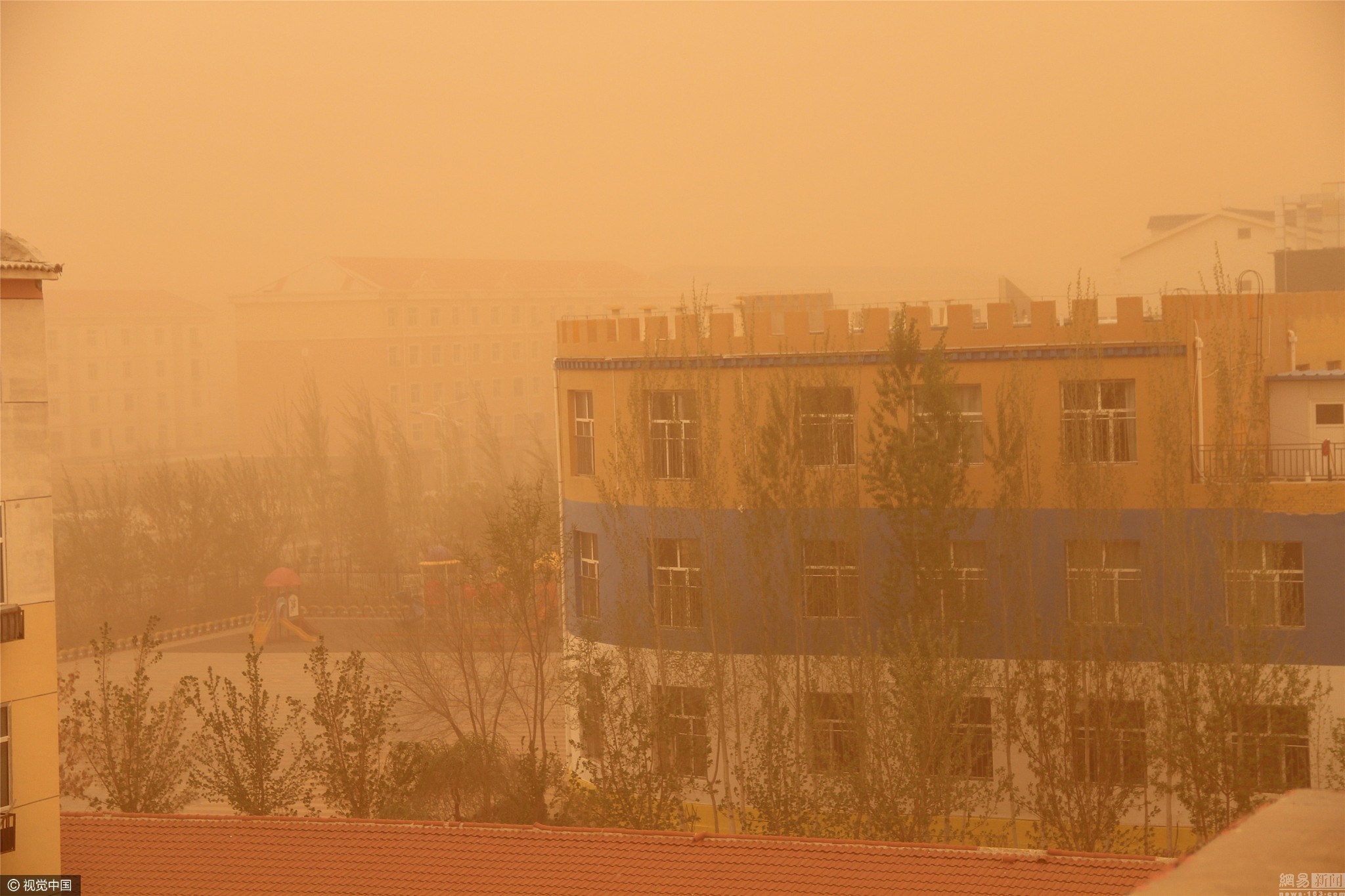}}
	\vspace{0.01cm}
\end{minipage}
\begin{minipage}[b]{.12\linewidth}
	\centering
	\centerline{\includegraphics[width=2.15cm]{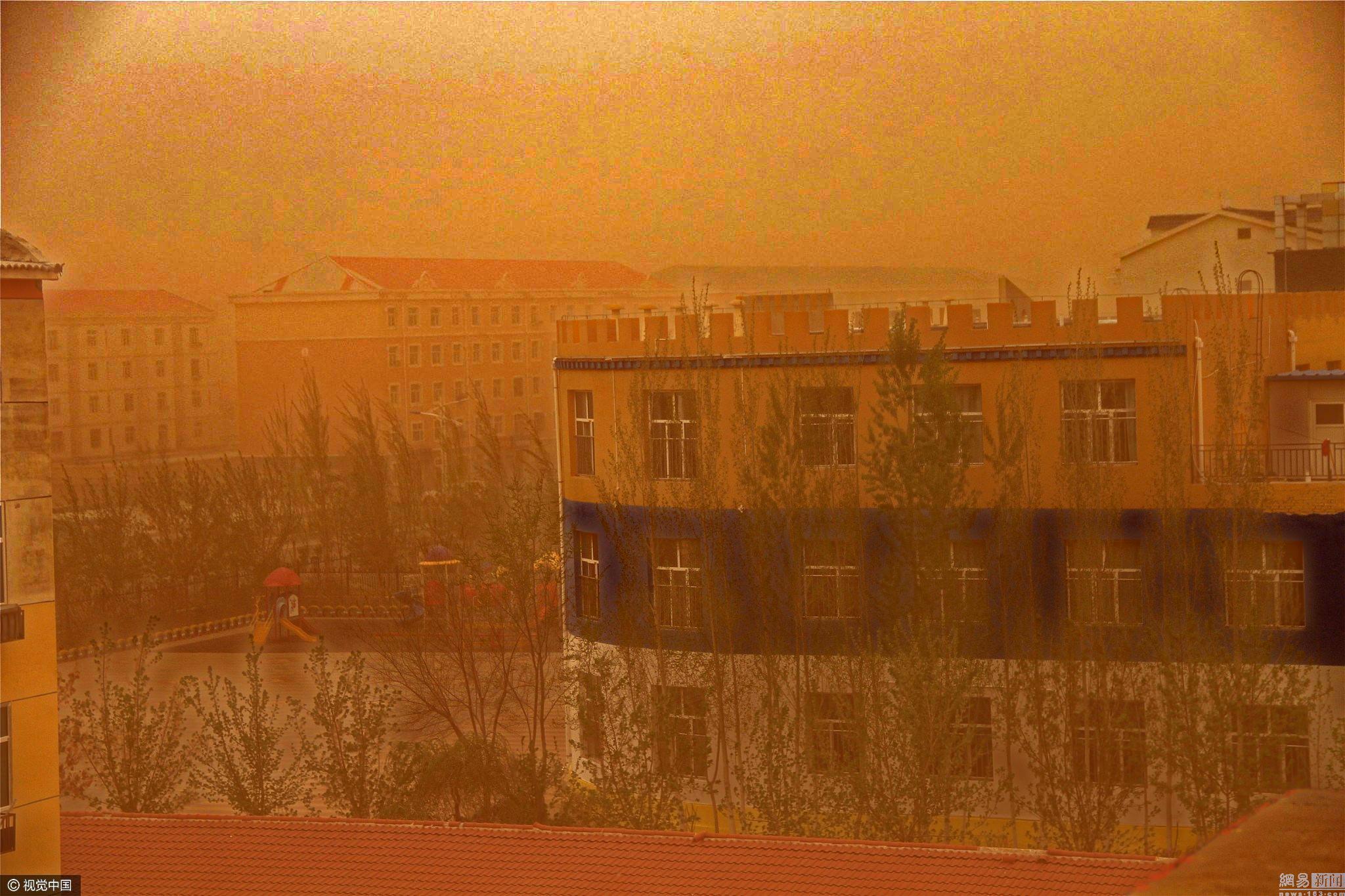}}
	\vspace{0.01cm}
\end{minipage}
\begin{minipage}[b]{0.12\linewidth}
	\centering
	\centerline{\includegraphics[width=2.15cm]{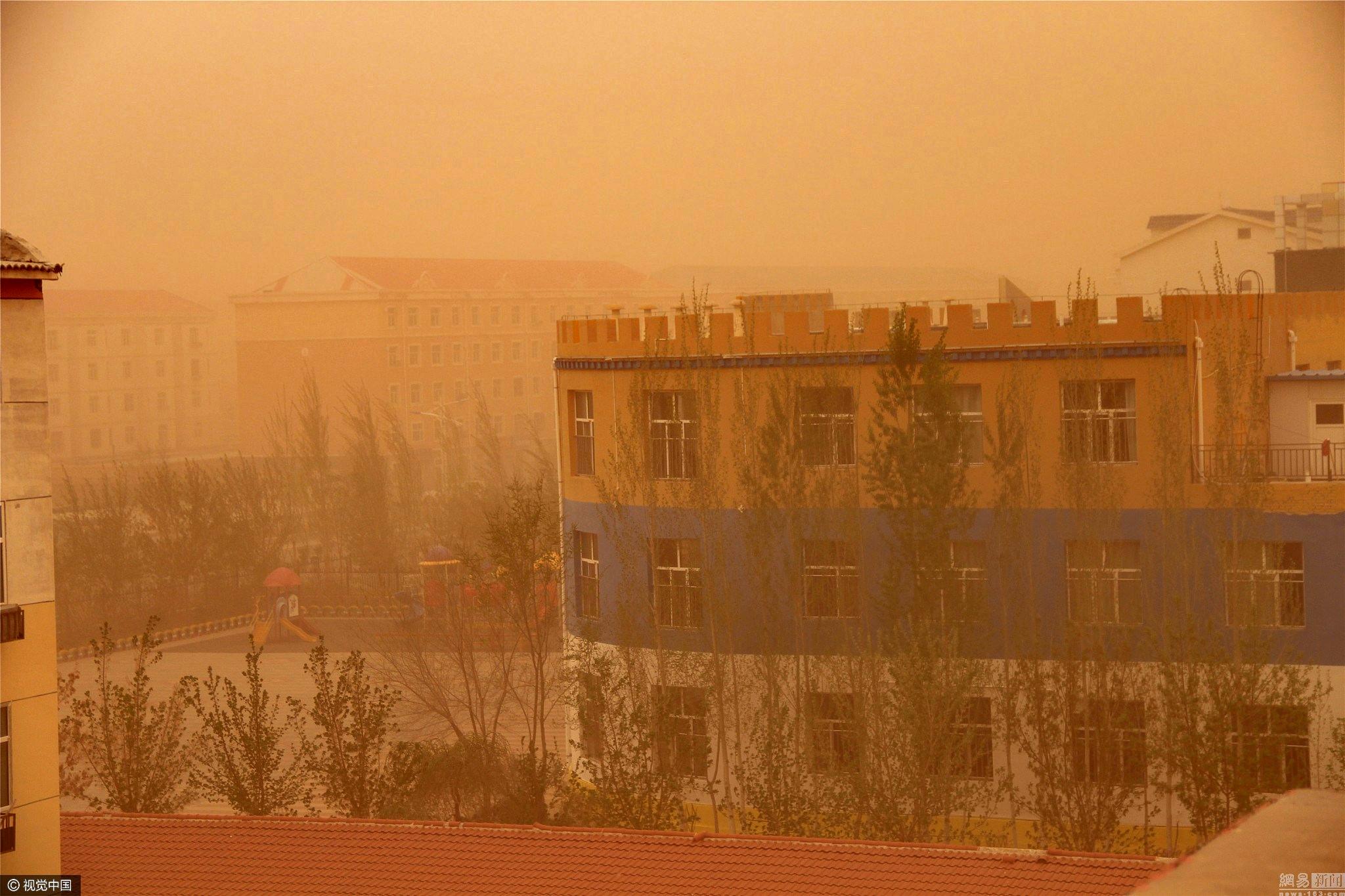}}
	\vspace{0.01cm}
\end{minipage}
\begin{minipage}[b]{0.12\linewidth}
	\centering
	\centerline{\includegraphics[width=2.15cm]{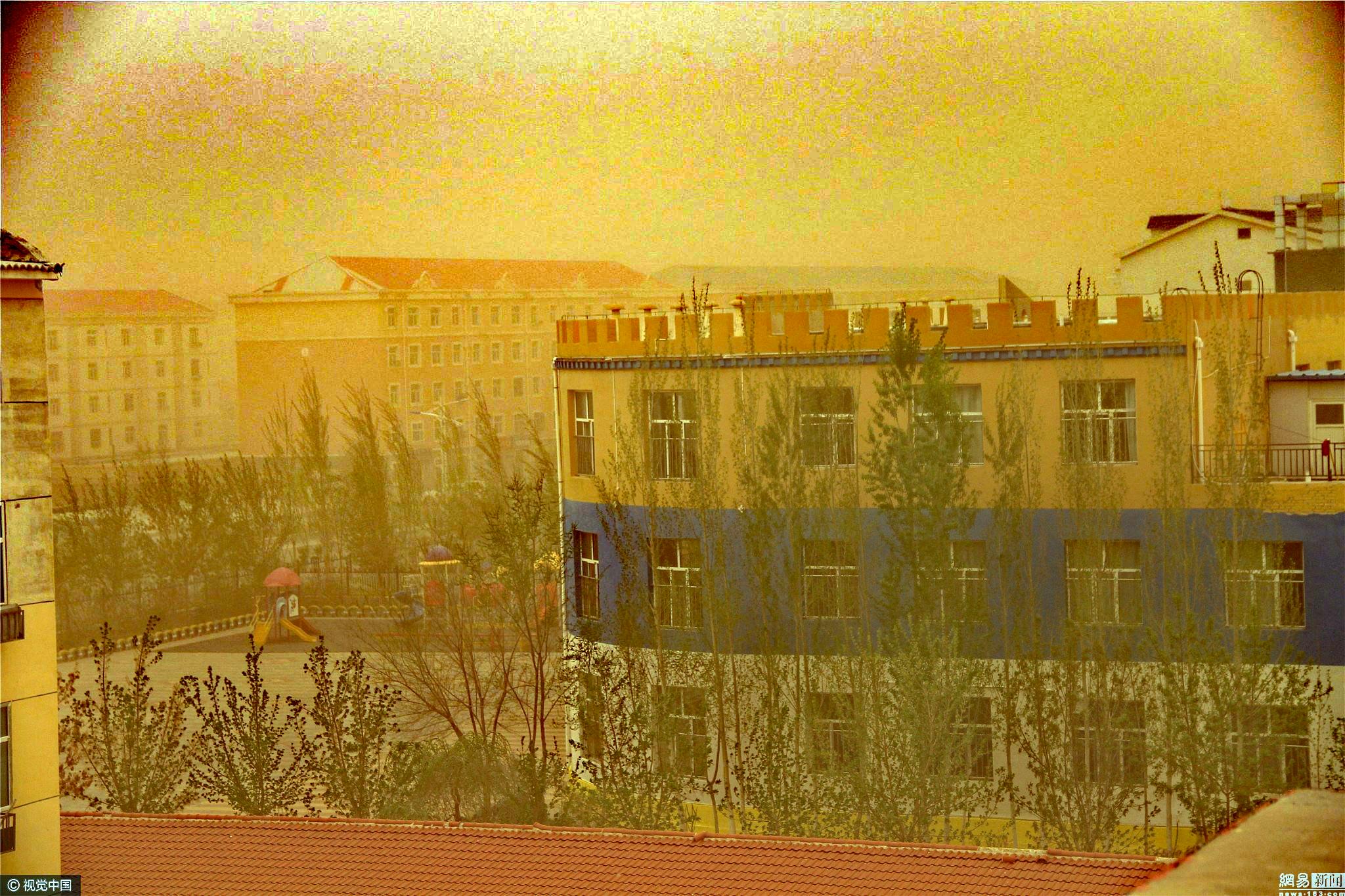}}
	\vspace{0.01cm}
\end{minipage}
\begin{minipage}[b]{0.12\linewidth}
	\centering
	\centerline{\includegraphics[width=2.15cm]{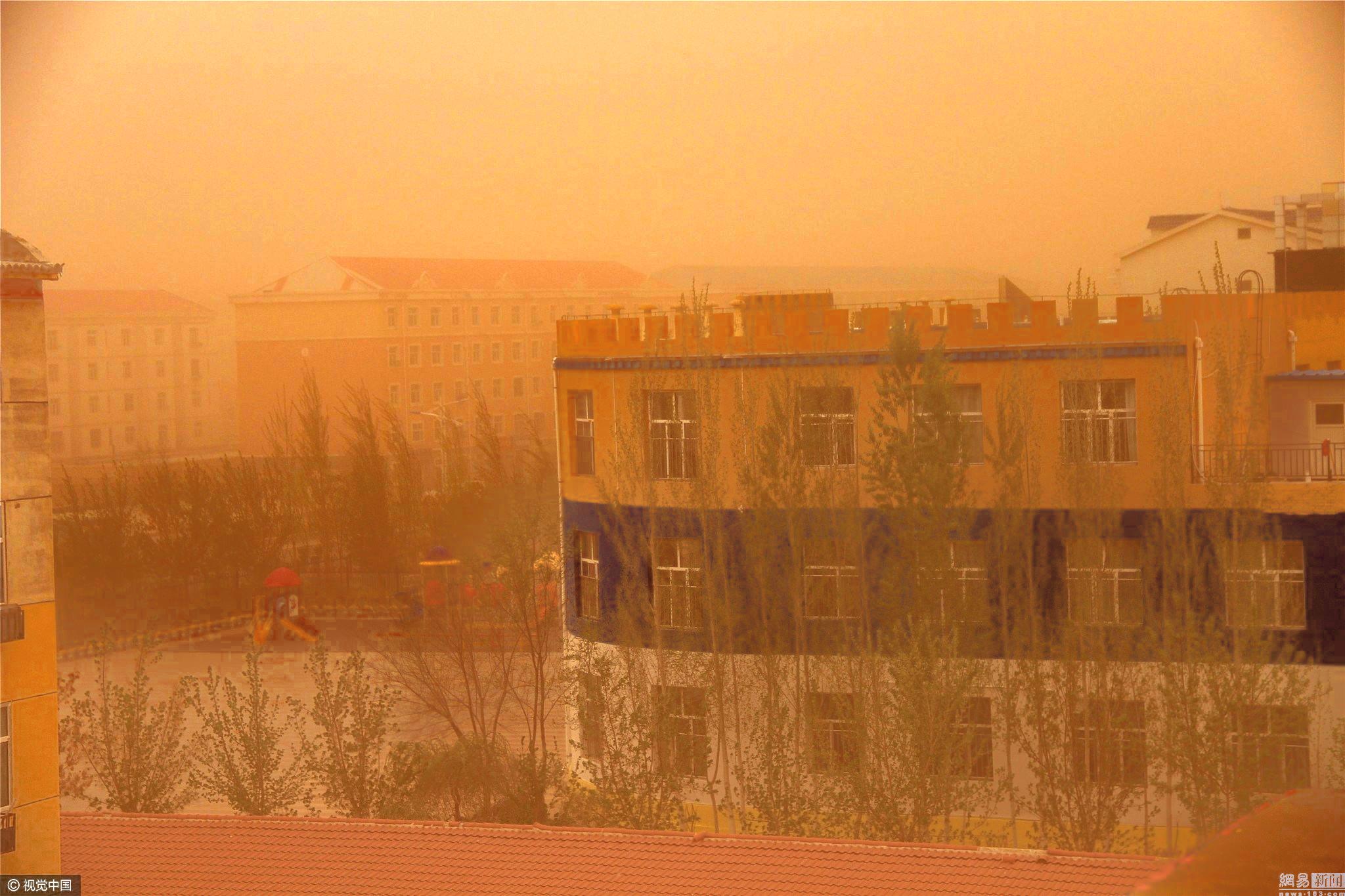}}
	\vspace{0.01cm}
\end{minipage}
\begin{minipage}[b]{0.12\linewidth}
	\centering
	\centerline{\includegraphics[width=2.15cm]{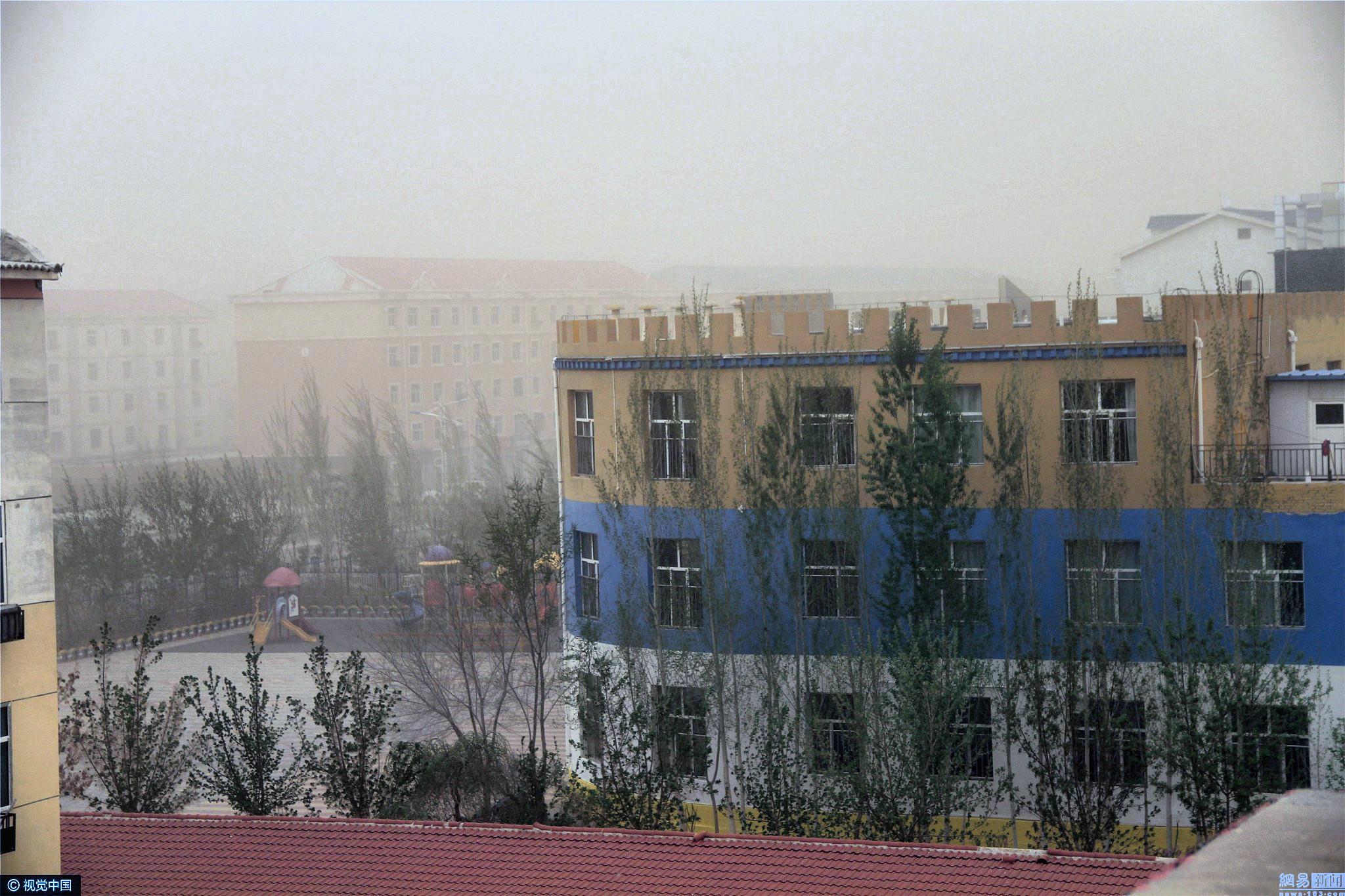}}
	\vspace{0.01cm}
\end{minipage}
\begin{minipage}[b]{.12\linewidth}
	\centering
	\centerline{\includegraphics[width=2.15cm]{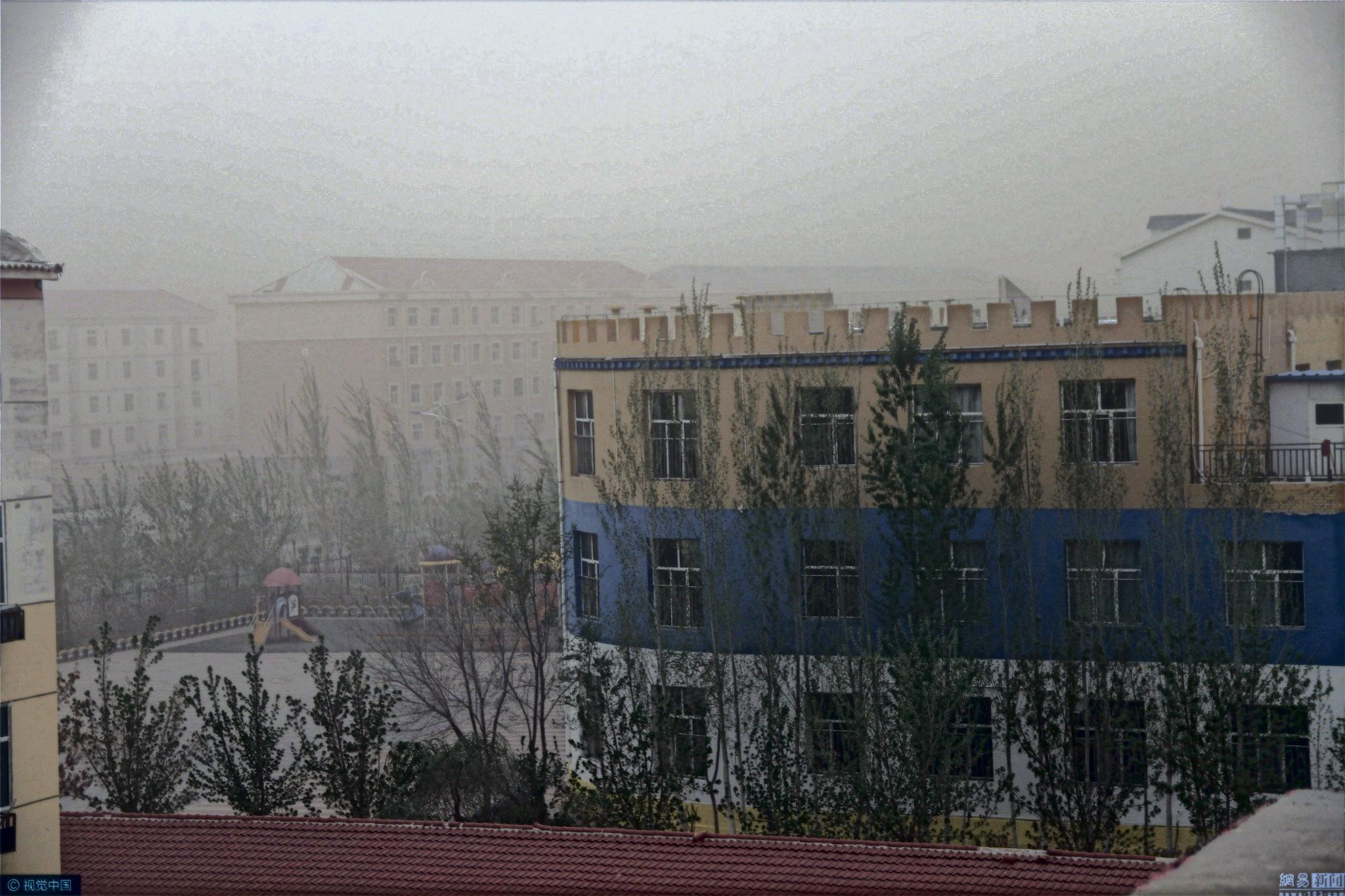}}
	\vspace{0.01cm}
\end{minipage}
\begin{minipage}[b]{0.12\linewidth}
	\centering
	\centerline{\includegraphics[width=2.15cm]{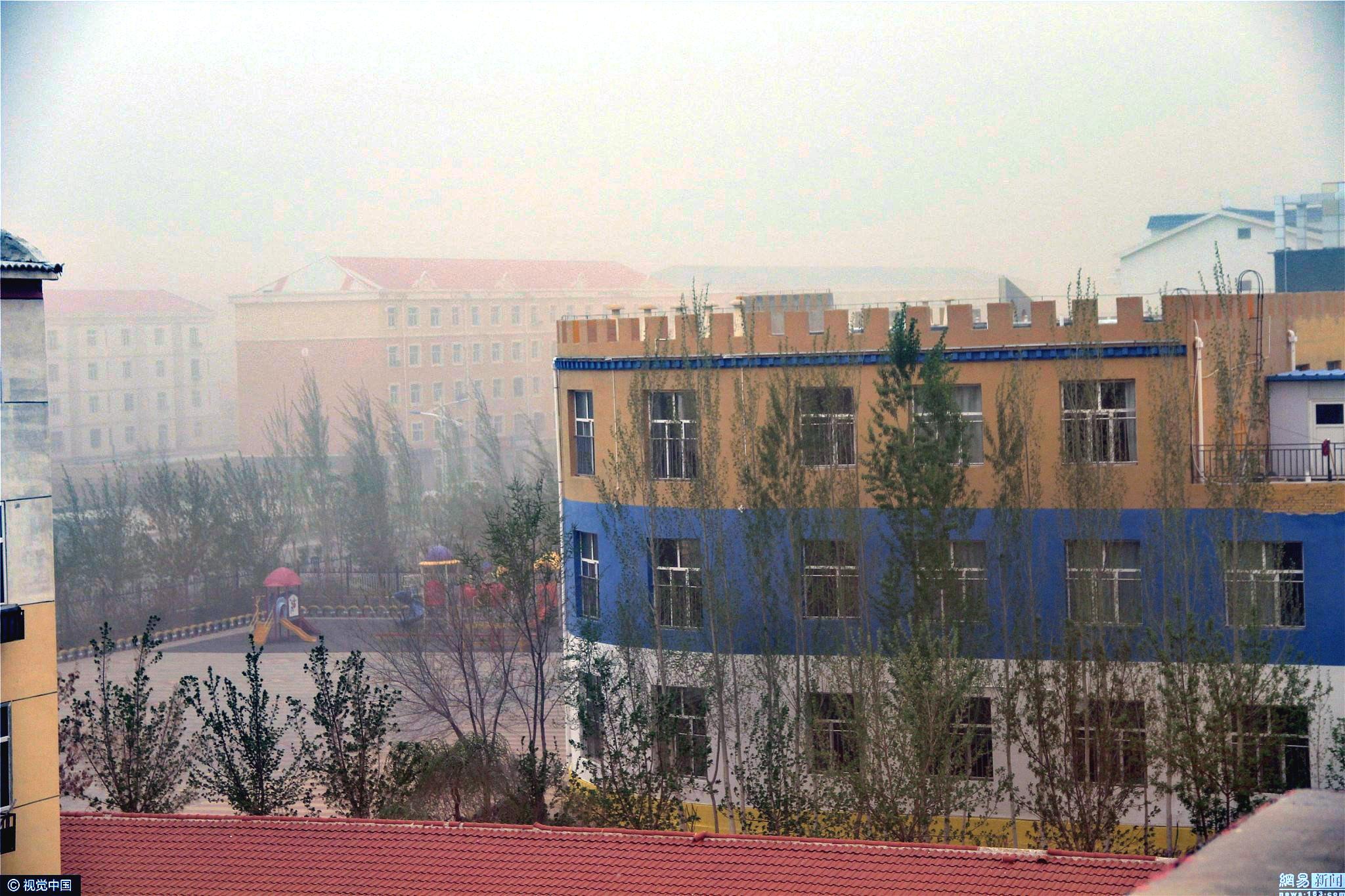}}
	\vspace{0.01cm}
\end{minipage}
\\
\begin{minipage}[b]{.12\linewidth}
	\centering
	\centerline{\includegraphics[width=2.15cm]{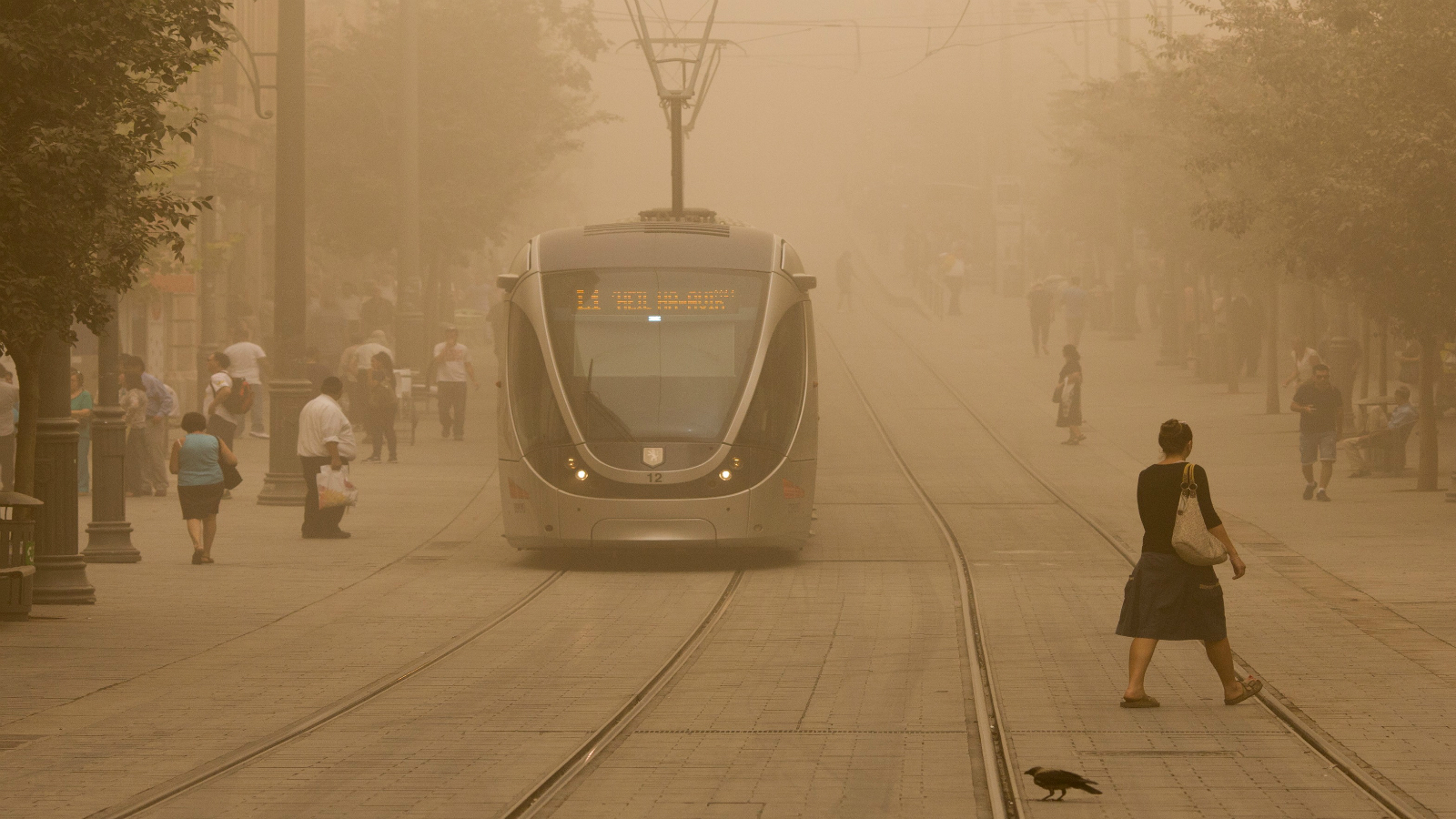}}
	\centerline{(a) Raw images}
\end{minipage}
\begin{minipage}[b]{.12\linewidth}
	\centering
	\centerline{\includegraphics[width=2.15cm]{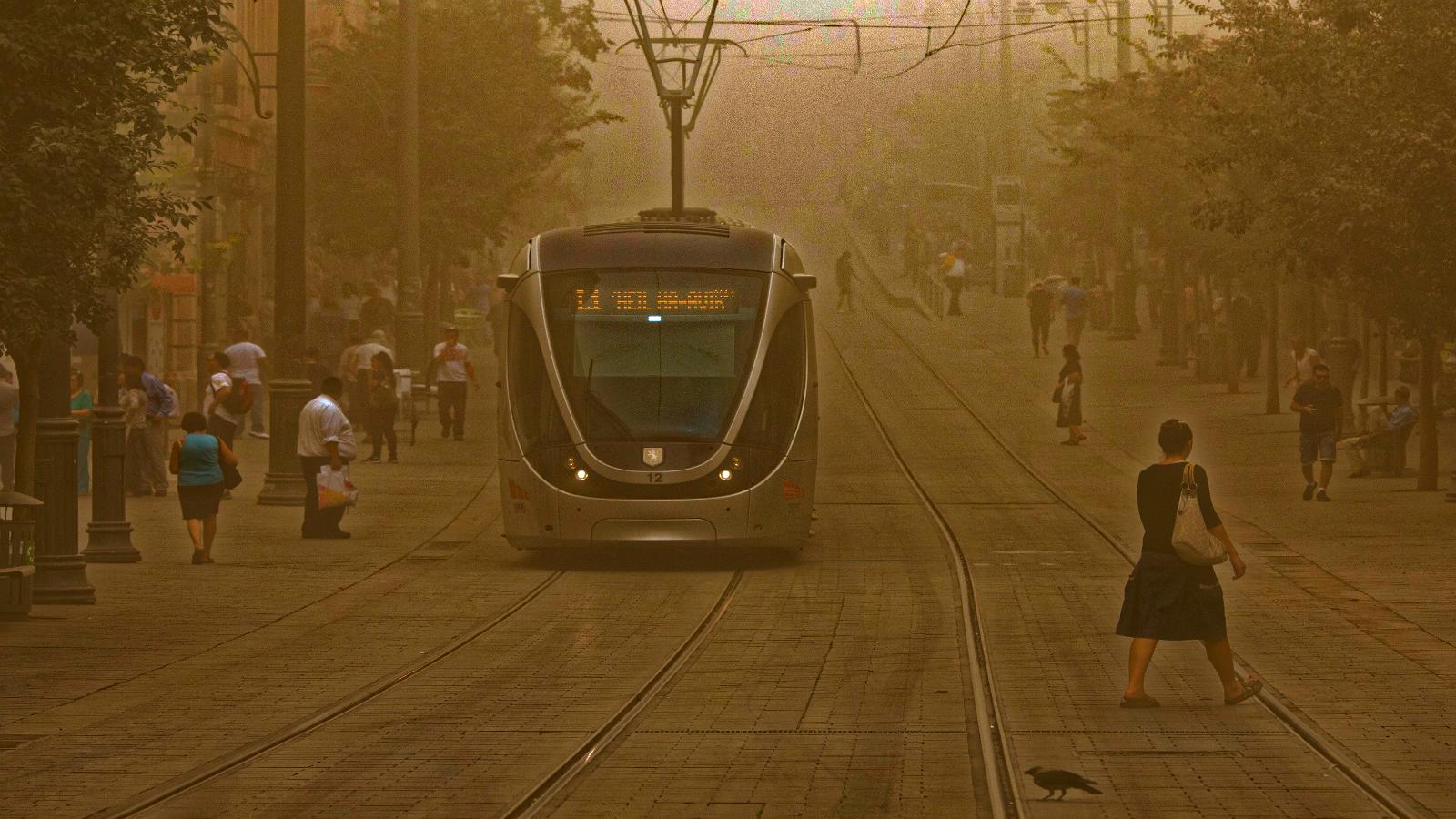}}
	\centerline{(b) DCP \cite{he2010single}}
\end{minipage}
\begin{minipage}[b]{0.12\linewidth}
	\centering
	\centerline{\includegraphics[width=2.15cm]{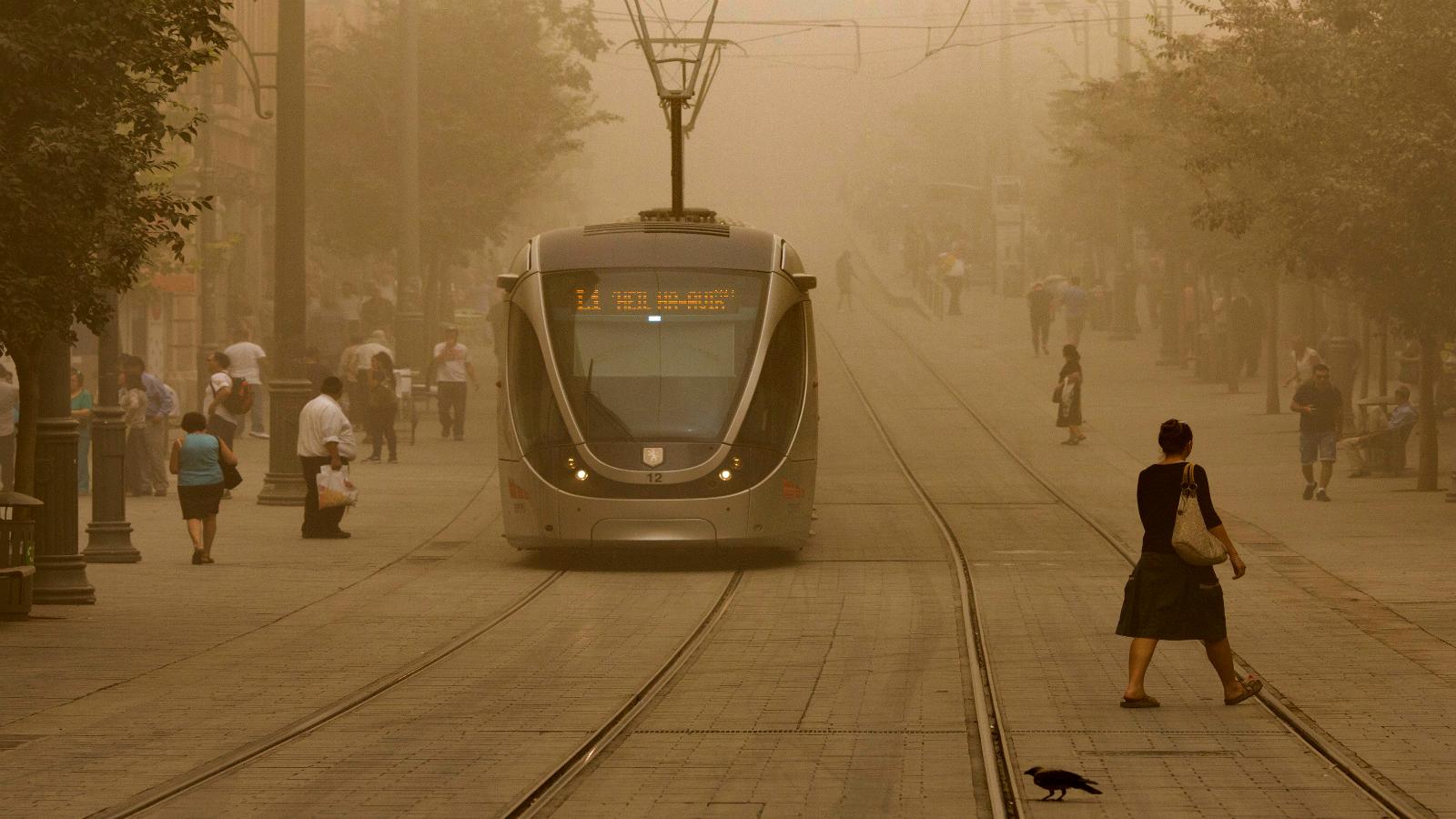}}
	\centerline{(c) MSCNN \cite{ren2016single}}
\end{minipage}
\begin{minipage}[b]{0.12\linewidth}
	\centering
	\centerline{\includegraphics[width=2.15cm]{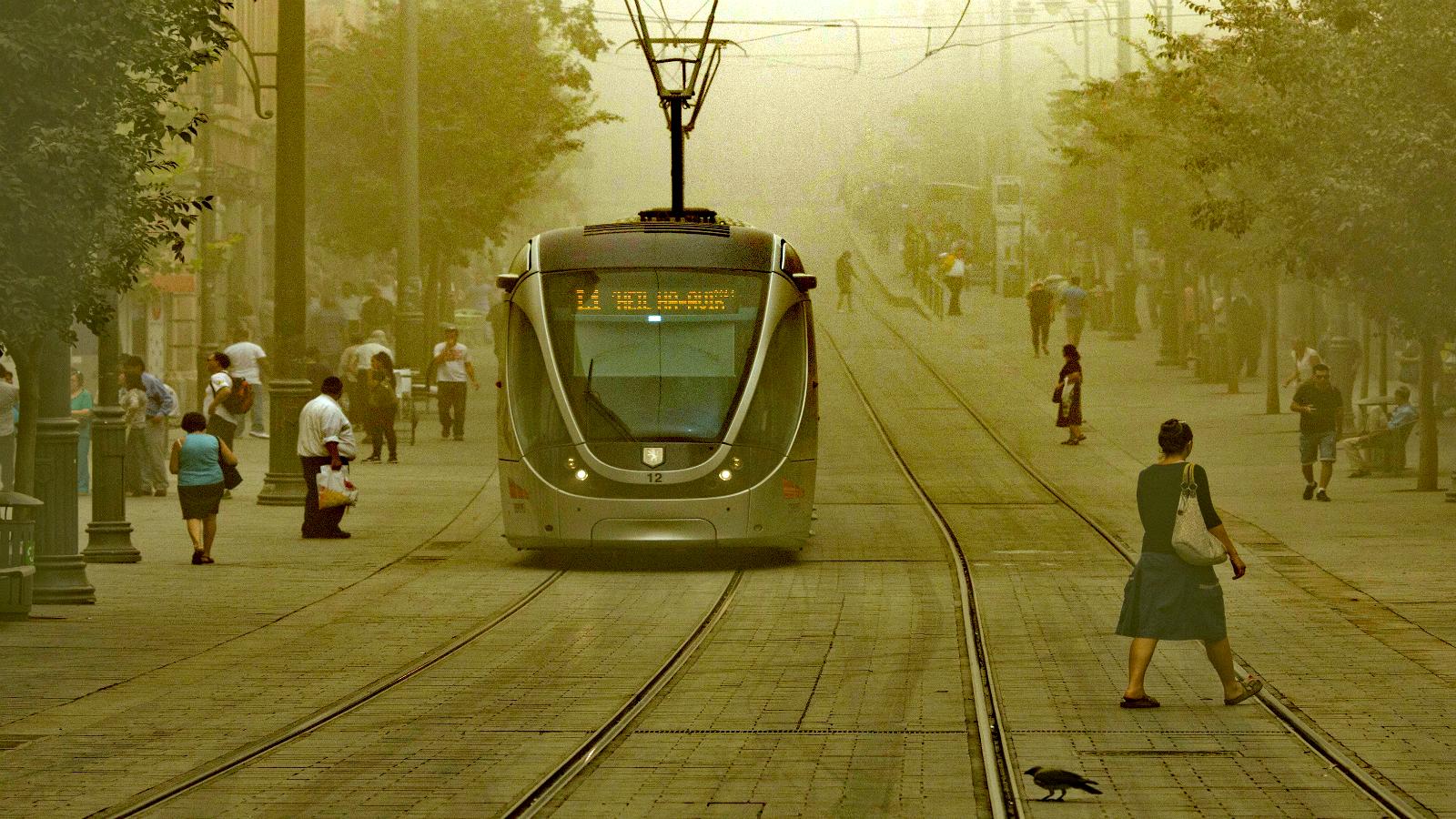}}
	\centerline{(d) Haze-Lines \cite{berman2017air}}
\end{minipage}
\begin{minipage}[b]{0.12\linewidth}
	\centering
	\centerline{\includegraphics[width=2.15cm]{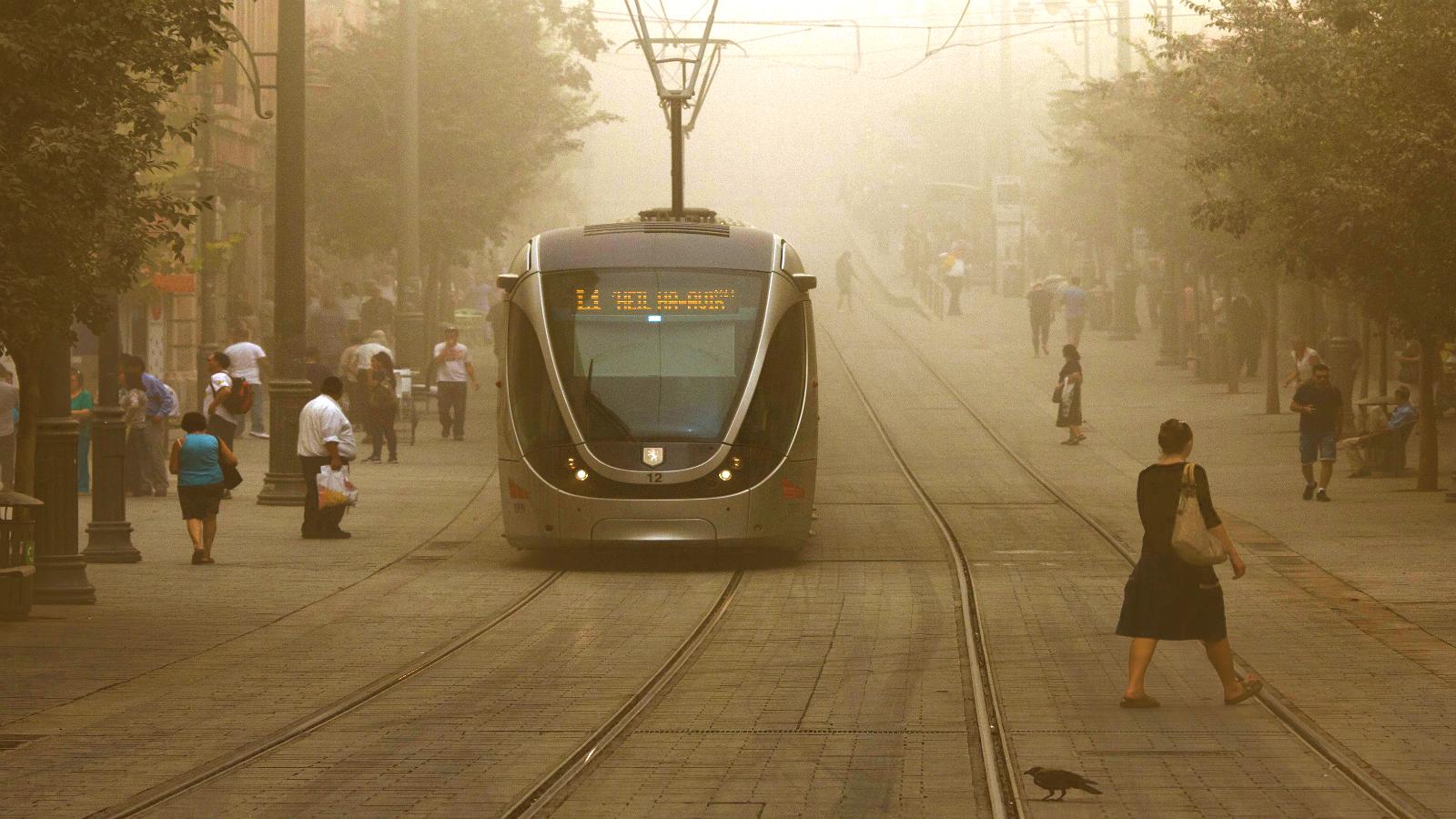}}
	\centerline{(e) LDCP \cite{zhu2018haze}}
\end{minipage}
\begin{minipage}[b]{0.12\linewidth}
	\centering
	\centerline{\includegraphics[width=2.15cm]{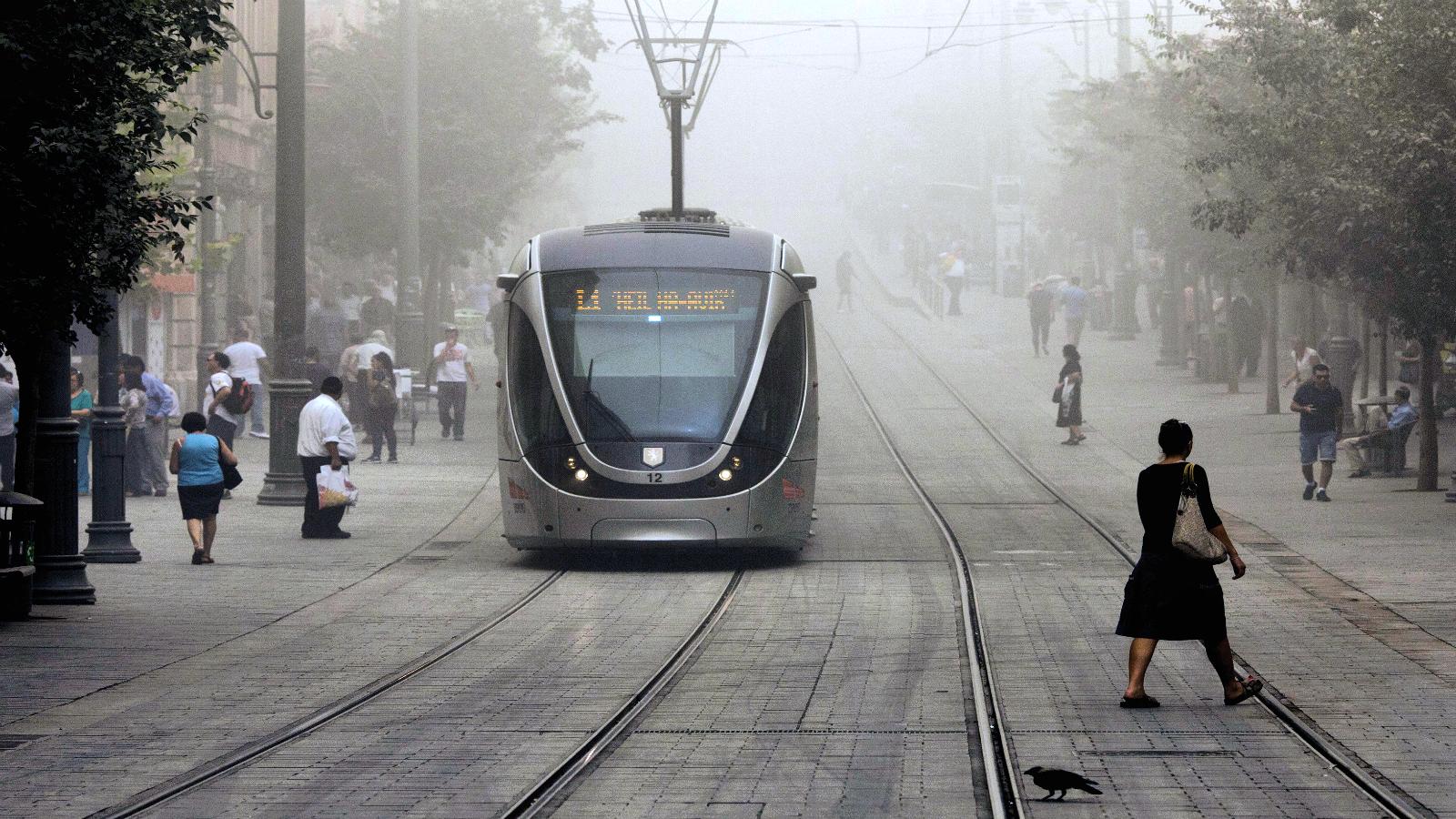}}
	\centerline{(f) Fusion \cite{fu2014fusion}}
\end{minipage}
\begin{minipage}[b]{0.12\linewidth}
	\centering
	\centerline{\includegraphics[width=2.15cm]{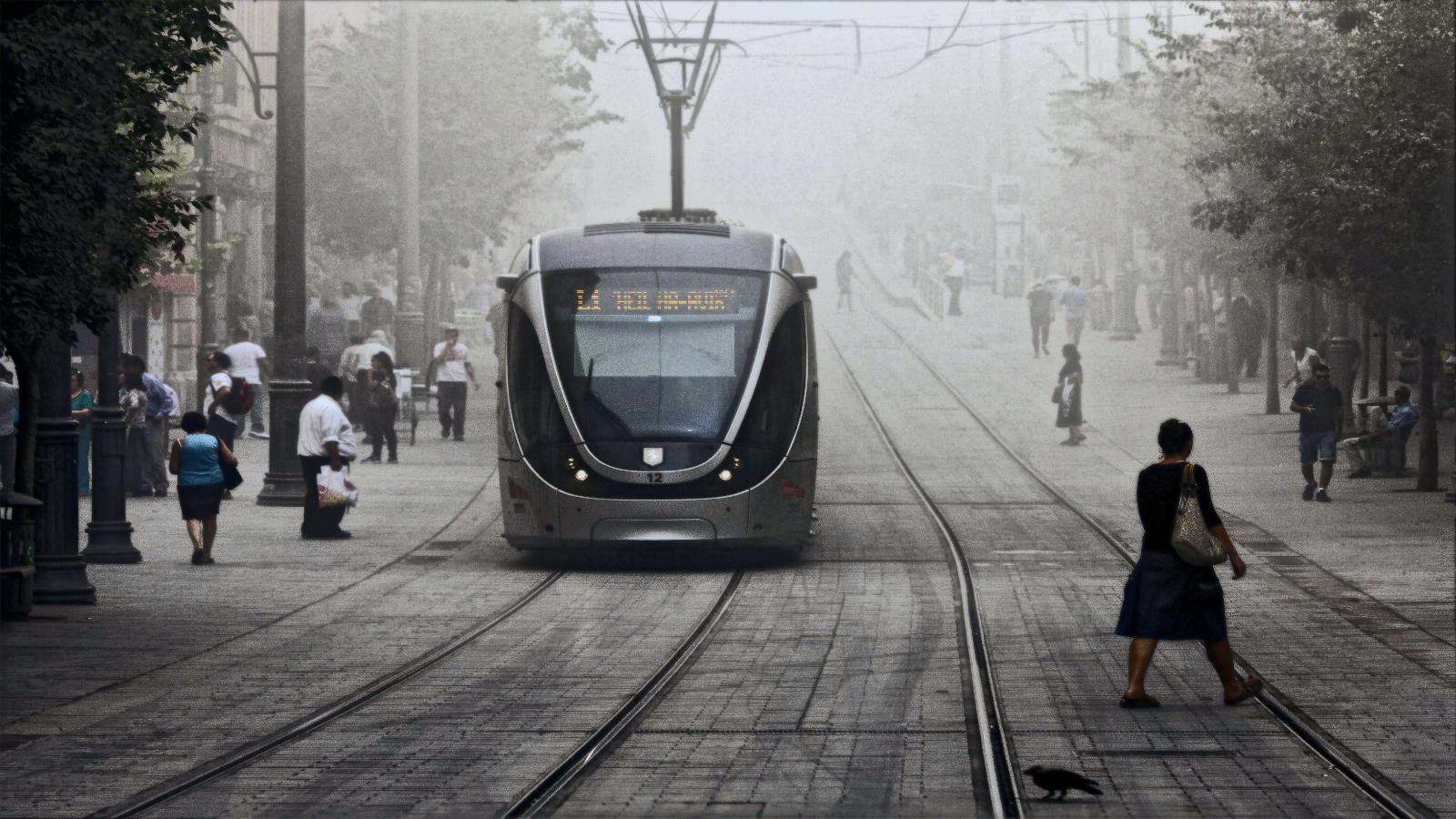}}
	\centerline{(g) Retinex \cite{fu2014retinex}}
\end{minipage}
\begin{minipage}[b]{0.12\linewidth}
	\centering
	\centerline{\includegraphics[width=2.15cm]{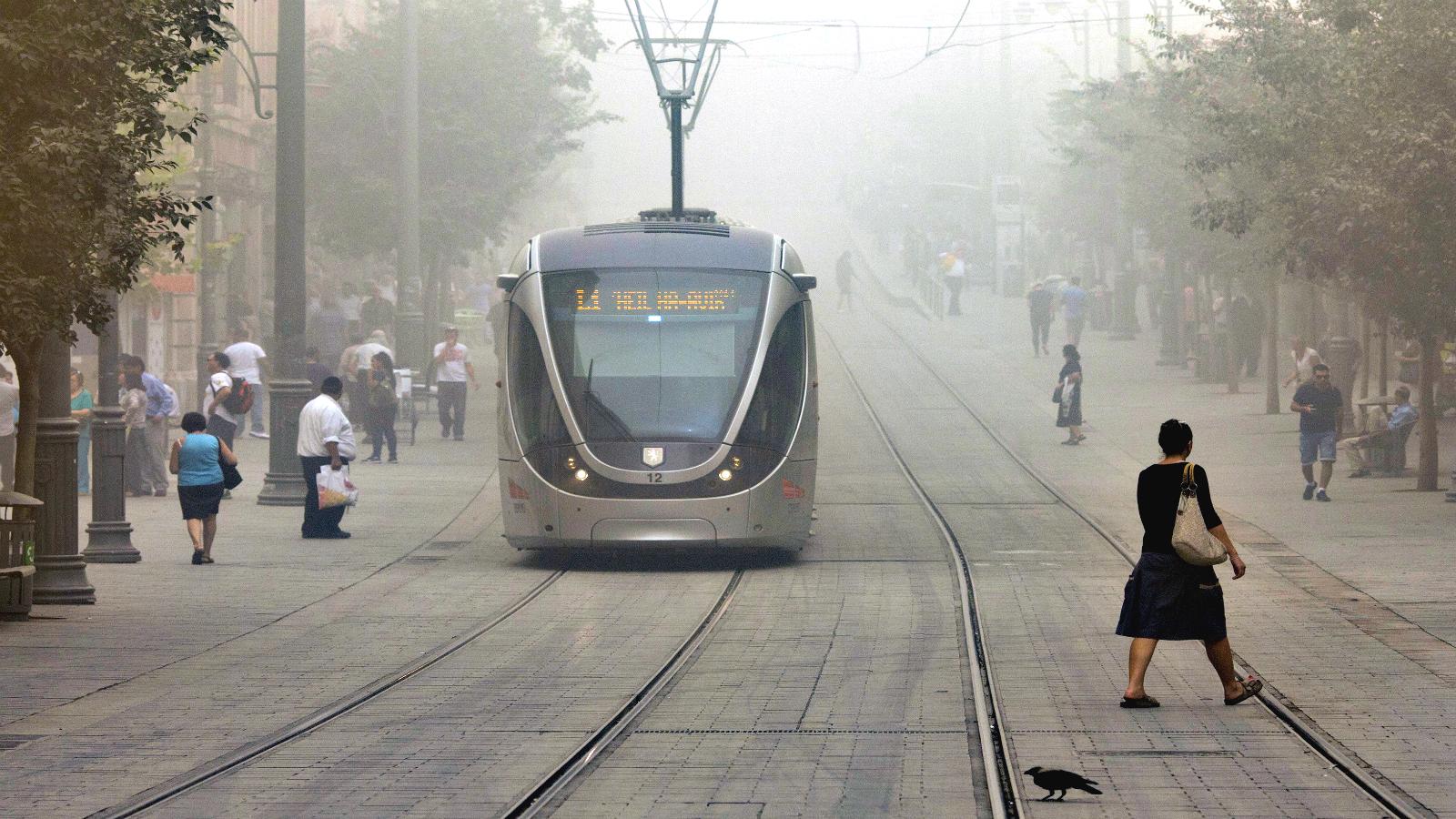}}
	\centerline{(h) Ours}
\end{minipage}
	\caption{Sandstorm image enhancement results obtained by different methods. (The images are best viewed in the full-screen mode.)}
	\label{fig:sandimages}
\end{figure*}

%%%%
\begin{figure*}[!t]
\begin{minipage}[b]{.162\linewidth}
\centering
\centerline{\includegraphics[width=2.87cm]{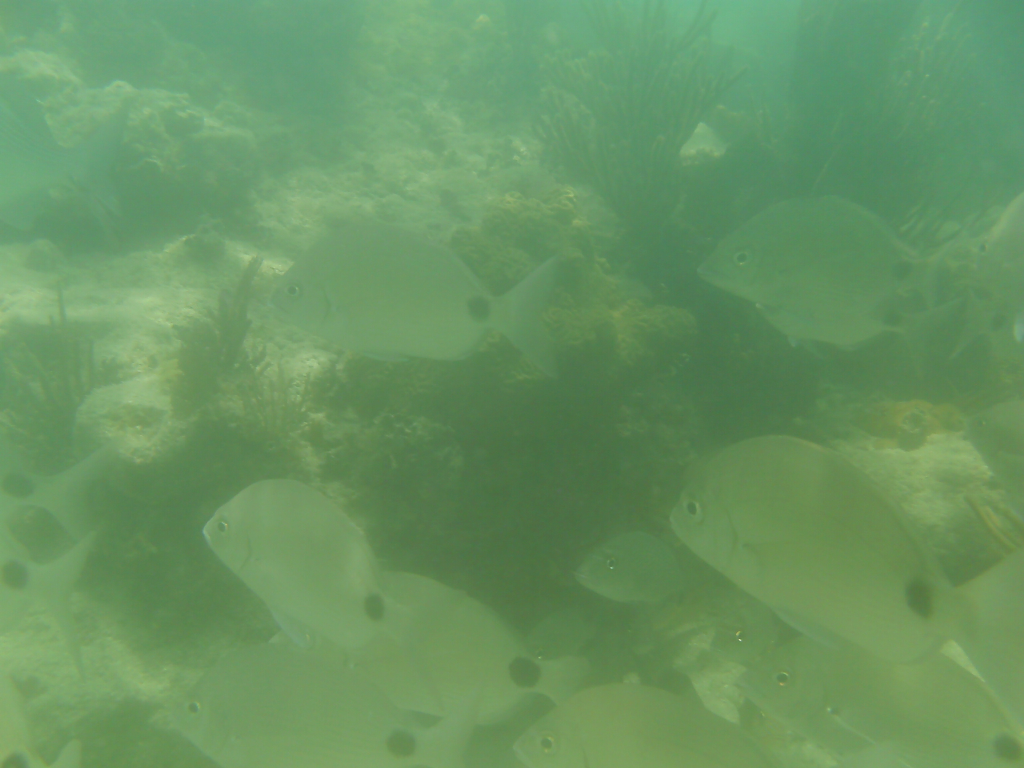}}
\vspace{0.01cm}
\end{minipage}
\begin{minipage}[b]{.162\linewidth}
\centering
\centerline{\includegraphics[width=2.87cm]{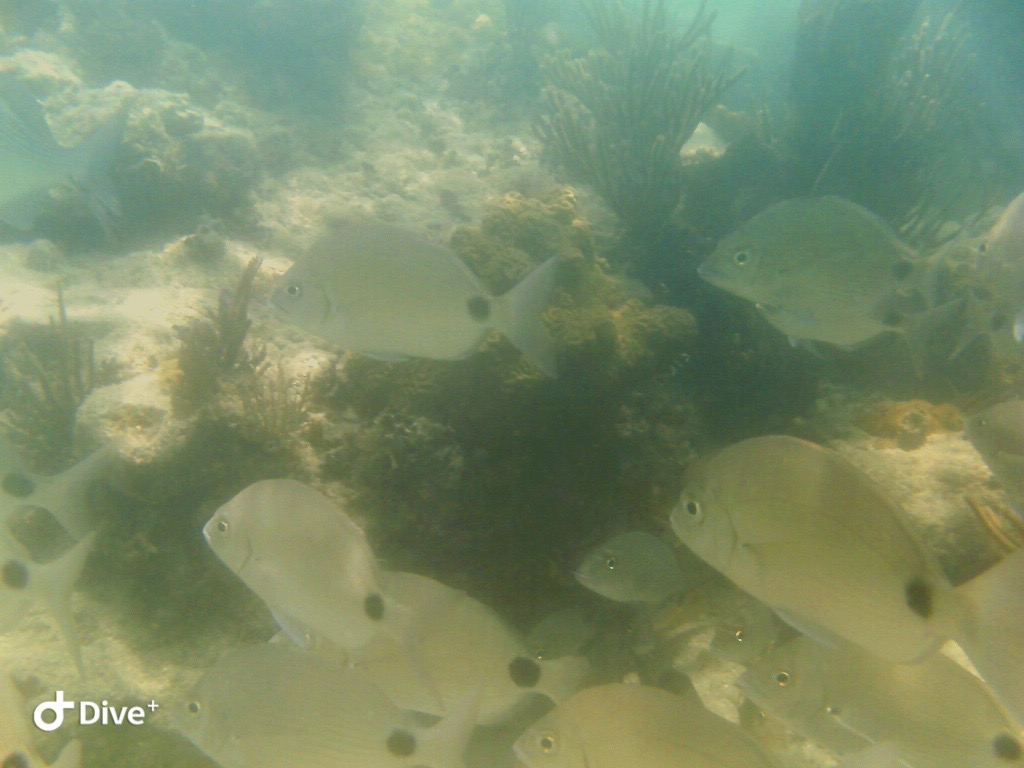}}
\vspace{0.01cm}
\end{minipage}
\begin{minipage}[b]{0.162\linewidth}
\centering
\centerline{\includegraphics[width=2.87cm]{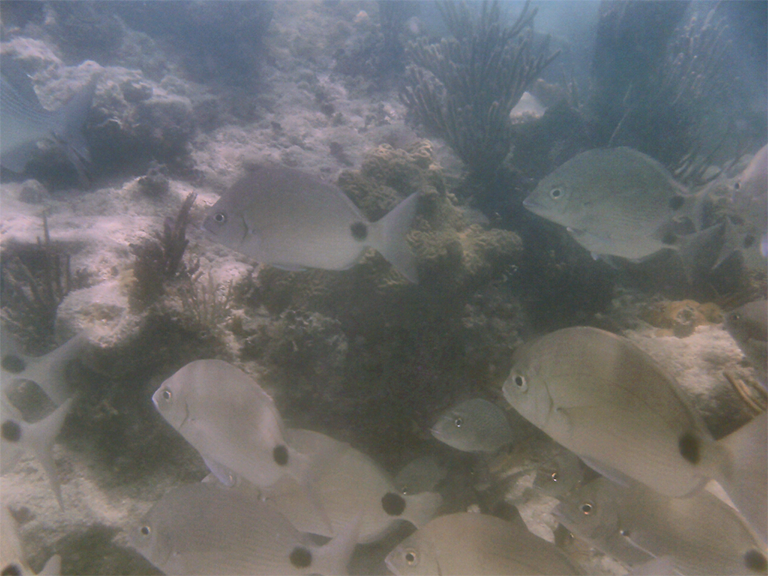}}
\vspace{0.01cm}
\end{minipage}
\begin{minipage}[b]{.162\linewidth}
\centering
\centerline{\includegraphics[width=2.87cm]{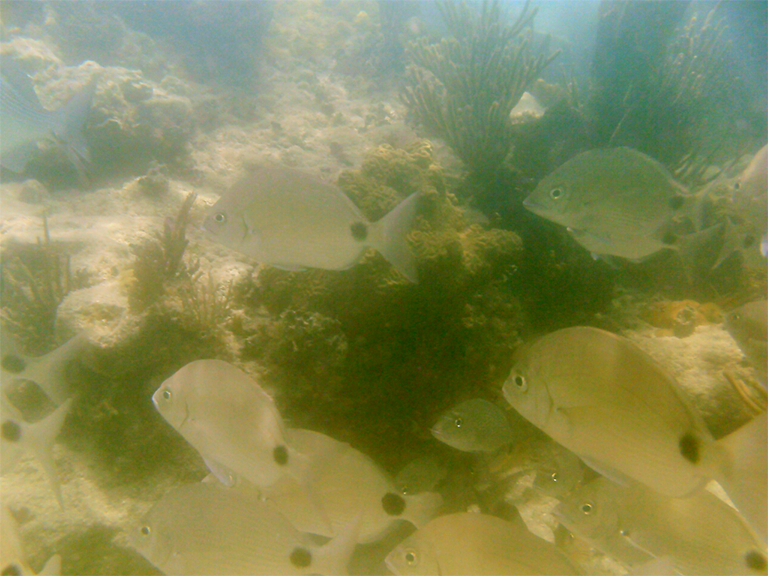}}
\vspace{0.01cm}
\end{minipage}
\begin{minipage}[b]{0.162\linewidth}
\centering
\centerline{\includegraphics[width=2.87cm]{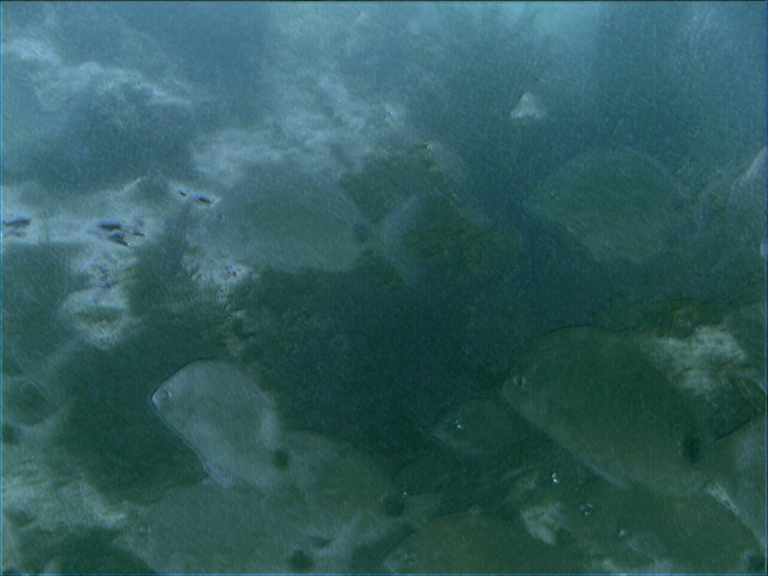}}
\vspace{0.01cm}
\end{minipage}
\begin{minipage}[b]{0.162\linewidth}
\centering
\centerline{\includegraphics[width=2.87cm]{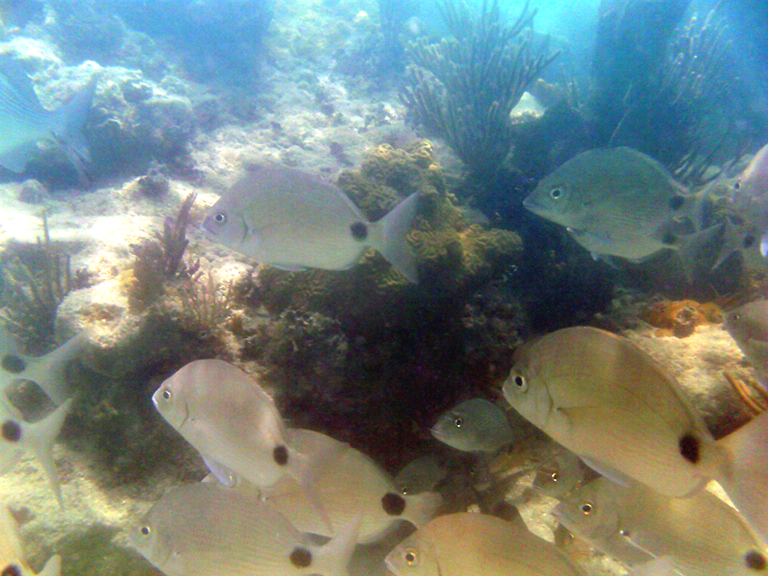}}
\vspace{0.01cm}
\end{minipage}
\\
\begin{minipage}[b]{.162\linewidth}
\centering
\centerline{\includegraphics[width=2.87cm]{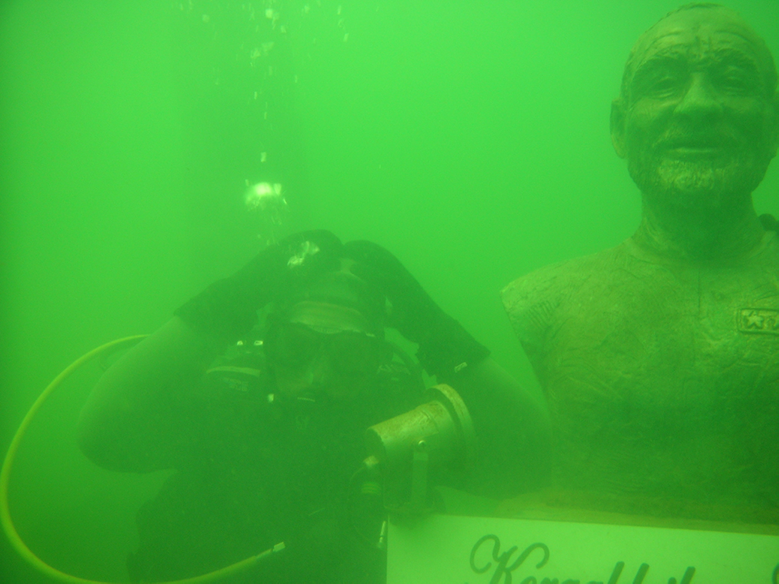}}
\centerline{\footnotesize{(a) Raw images}}\medskip
\vspace{0.001cm}
\end{minipage}
\begin{minipage}[b]{.162\linewidth}
\centering
\centerline{\includegraphics[width=2.87cm]{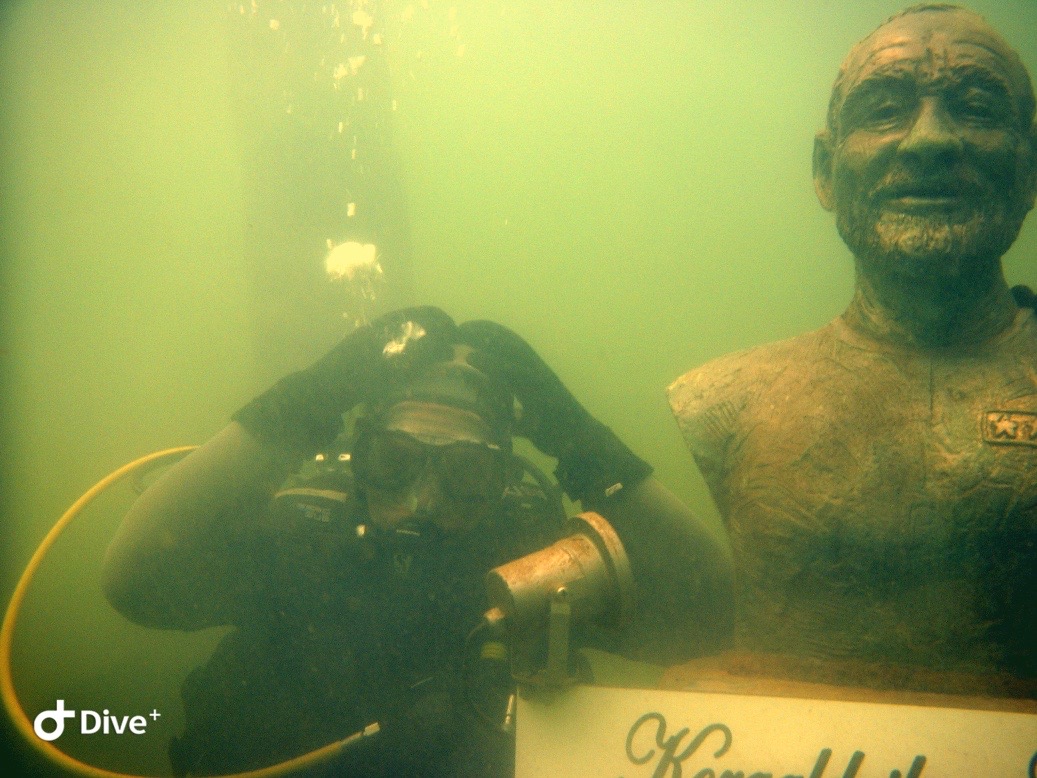}}
\centerline{\footnotesize{(b) Dive+}}\medskip
\vspace{0.01cm}
\end{minipage}
\begin{minipage}[b]{0.162\linewidth}
\centering
\centerline{\includegraphics[width=2.87cm]{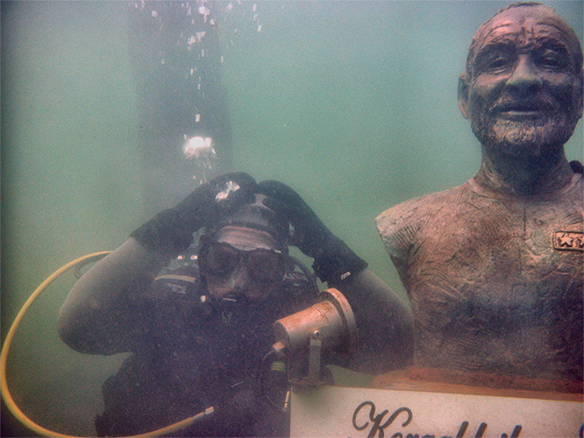}}
\centerline{\footnotesize{(c) Two-step \cite{fu2017two}} }\medskip
\vspace{0.001cm}
\end{minipage}
\begin{minipage}[b]{.162\linewidth}
\centering
\centerline{\includegraphics[width=2.87cm]{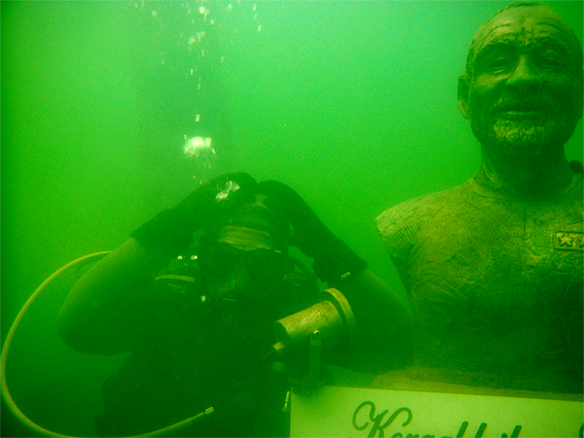}}
\centerline{\footnotesize{(d) Blurriness \cite{peng2017underwater}}}\medskip
\vspace{0.01cm}
\end{minipage}
\begin{minipage}[b]{.162\linewidth}
\centering
\centerline{\includegraphics[width=2.87cm]{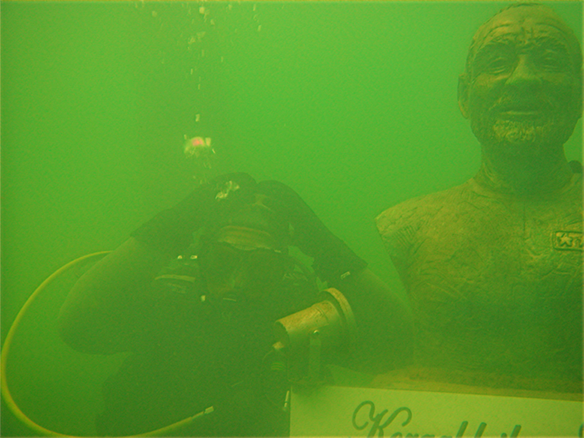}}
\centerline{\footnotesize{(e) UWCNN \cite{li2020underwater} }}\medskip
\vspace{0.01cm}
\end{minipage}
\begin{minipage}[b]{0.162\linewidth}
\centering
\centerline{\includegraphics[width=2.87cm]{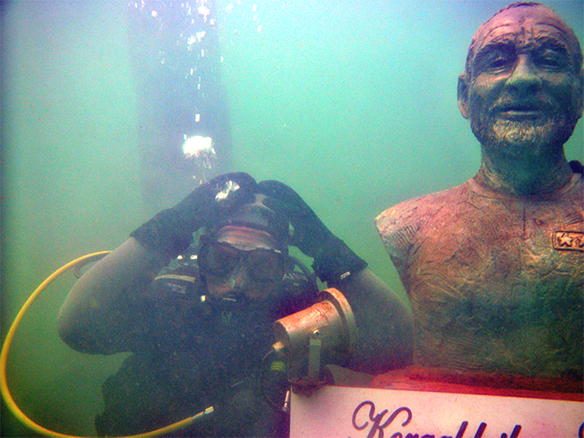}}
\centerline{\footnotesize{(f) Ours } }\medskip
\vspace{0.01cm}
\end{minipage}
\caption{Enhanced results obtained by different methods. (The images are best viewed in the full-screen mode.)}
\label{fig:underwatercmp}
\end{figure*}

%%%
 
\begin{figure*}[!t]
\begin{minipage}[b]{.162\linewidth}
\centering
\centerline{\includegraphics[width=2.87cm]{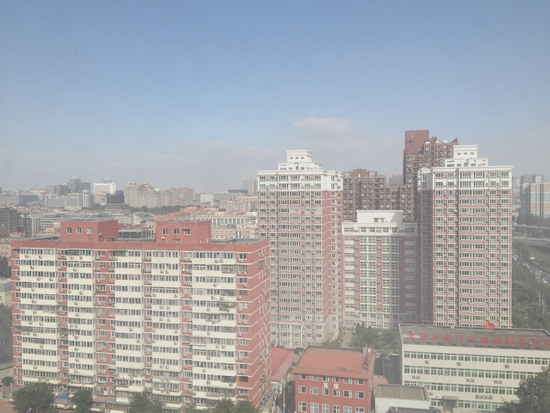}}
\vspace{0.01cm}
\end{minipage}
\begin{minipage}[b]{.162\linewidth}
\centering
\centerline{\includegraphics[width=2.87cm]{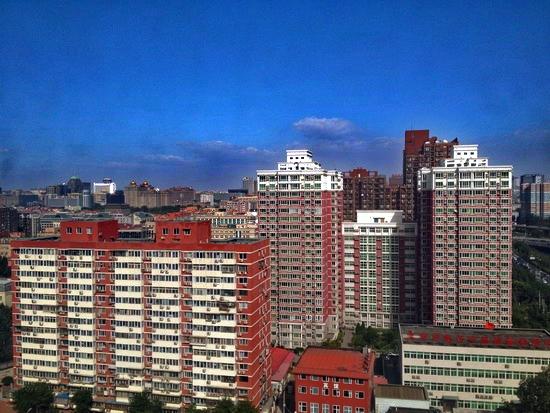}}
\vspace{0.01cm}
\end{minipage}
\begin{minipage}[b]{0.162\linewidth}
\centering
\centerline{\includegraphics[width=2.87cm]{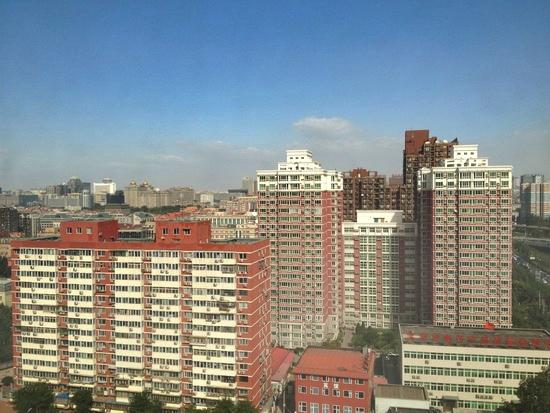}}
\vspace{0.01cm}
\end{minipage}
\begin{minipage}[b]{.162\linewidth}
\centering
\centerline{\includegraphics[width=2.87cm]{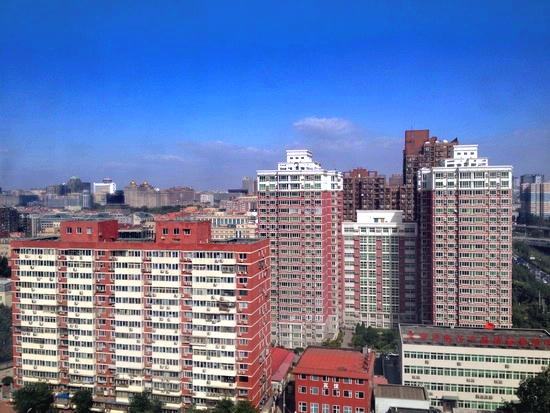}}
\vspace{0.01cm}
\end{minipage}
\begin{minipage}[b]{0.162\linewidth}
\centering
\centerline{\includegraphics[width=2.87cm]{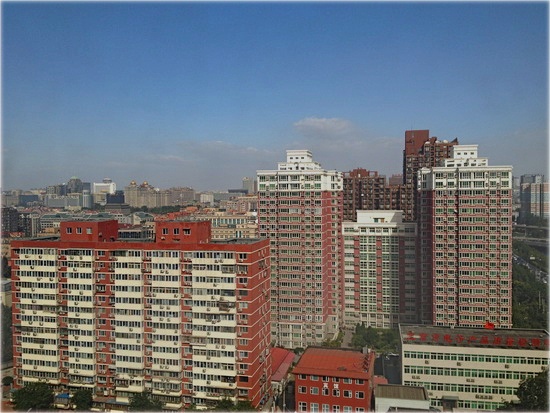}}
\vspace{0.01cm}
\end{minipage}
\begin{minipage}[b]{0.162\linewidth}
\centering
\centerline{\includegraphics[width=2.87cm]{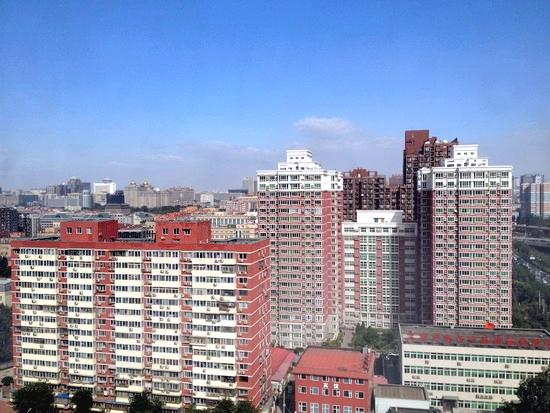}}
\vspace{0.01cm}
\end{minipage}
\\
\begin{minipage}[b]{.162\linewidth}
\centering
\centerline{\includegraphics[width=2.87cm]{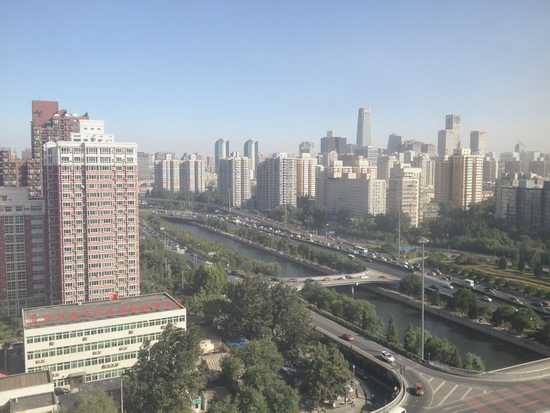}}
\centerline{\footnotesize{(a) Raw images}}\medskip
\end{minipage}
\begin{minipage}[b]{.162\linewidth}
\centering
\centerline{\includegraphics[width=2.87cm]{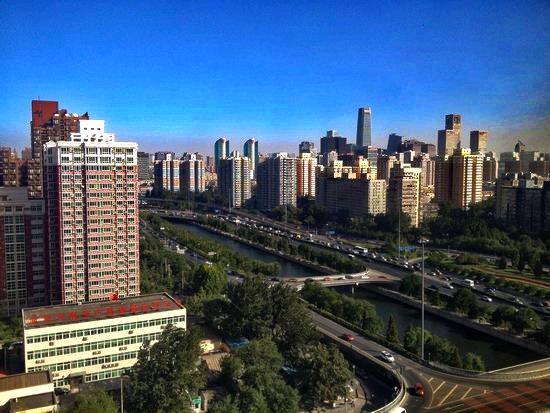}}
\centerline{\footnotesize{(b) DCP \cite{he2010single}}}\medskip
\end{minipage}
\begin{minipage}[b]{0.162\linewidth}
\centering
\centerline{\includegraphics[width=2.87cm]{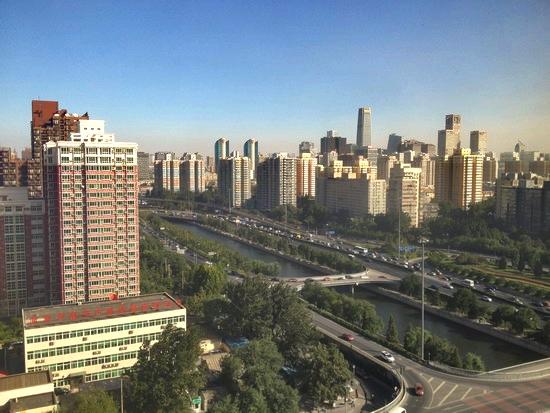}}
\centerline{\footnotesize{(c) MSCNN \cite{ren2016single}}}\medskip
\end{minipage}
\begin{minipage}[b]{.162\linewidth}
\centering
\centerline{\includegraphics[width=2.87cm]{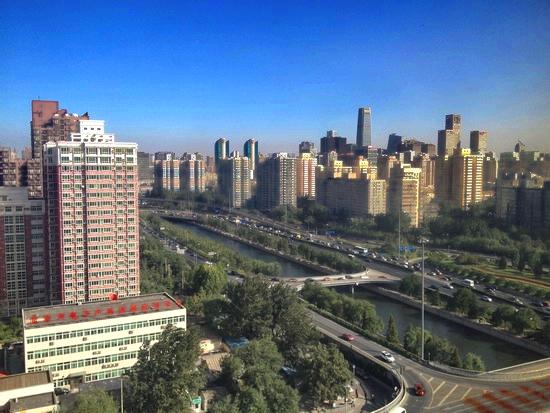}}
\centerline{\footnotesize{(d) LDCP \cite{zhu2018haze}}}\medskip
\end{minipage}
\begin{minipage}[b]{.162\linewidth}
\centering
\centerline{\includegraphics[width=2.87cm]{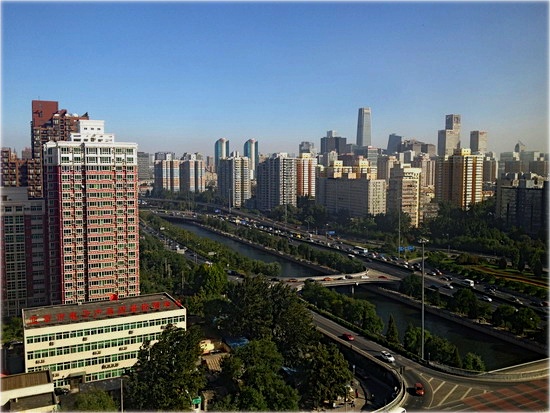}}
\centerline{\footnotesize{(e) AODNet \cite{li2017aod}}}\medskip
\end{minipage}
\begin{minipage}[b]{0.162\linewidth}
\centering
\centerline{\includegraphics[width=2.87cm]{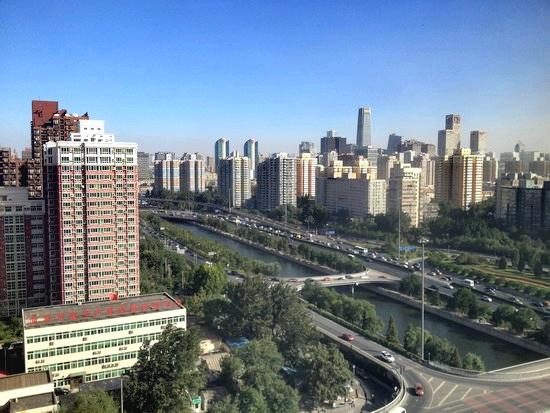}}
\centerline{\footnotesize{(f) Ours}  }\medskip
\end{minipage}
\caption{Dehazing results obtained by different methods. (The images are best viewed in the full-screen mode.) }
\label{fig:dehazecmp}
\end{figure*}

\subsection{Unified spectrum}
Eqs. (\ref{eq:IMF}) and (\ref{rad_inten}) show that, for the scenes in a long distance apart from the camera, due to the effect of the ambient transmission, the obtained image $\mathbf{I}$ is mainly the irradiation of the ambient light source. For the near scenes, the imaging radiation contains a relatively small amount of ambient light source irradiation. Although the radiation intensity of the scene is different in imaging, they reflect the same radiation spectrum, which we name the unified spectrum. Figure \ref{fig:flowchart} visually demonstrates what is the unified spectrum.

For the observed intensity $\mathbf{I}$, the unified radiance $\mathbf{S}_u \in \mathbb{R}^{1\times 3}$ is given by the following formula:
\begin{equation}\label{comp_Iu}
	\min_{\mathbf{S}_u^c}\left\|\mathbf{I}^{c}-\mathbf{S}_{u}^{c}\right\|_{F}^{2},
\end{equation}
where $\|\cdot\|_F$ denotes the Frobenius norm and $c \in \{R, G, B\}$ is the color channel of $\mathbf{I}$. The closed-form solution of (\ref{comp_Iu}) can be directly computed by mean:
\begin{equation}\label{Iu}
	\mathbf{S}_{u}^{c}=\frac{1}{|\Omega|} \sum_{x \in \Omega} \mathbf{I}^{c}(x), \quad c \in\{R, G, B\}.
\end{equation}
The unified radiance reflects the homogenous ambient light. To describe its essential nature, we normalize the unified radiance via the following formula:
\begin{equation}\label{eq:unified_spectrum}
	\mathbf{S}_{nu} = \frac{\mathbf{S}_u}{\|\mathbf{S}_u\|_1},
\end{equation}
where $\mathbf{S}_{nu}$ is called the unified spectrum. It describes the spectral characteristics of the ambient light source. We claim that the unified spectrum $\mathbf{S}_{nu}$ is a good approximation of  transmission basis $\mathbf{I}_u$. That is to say, $\mathbf{S}_{nu}$ approximates the common direction for the transmission.  Then, here comes  a new way for estimating the transmission $\tilde{t}$ and ambient light source $\mathbf{A}$ (details in the next section). 

\section{Scene Recovery with Rank-One Prior}
\subsection{Estimation of the transmission}
To differentiate from the existing haze removal methods which rely overly on the local contrast of degraded images, the new strategy focuses on the optical principle of degraded images and is physically valid.
\par
\begin{equation}\label{estimatedtransformula}
\tilde{t}(x)=\langle \mathbf{I}(x), \mathbf{S}_{nu} \rangle\cdot {\mathbf{S}_{nu}}.
\end{equation}
This formula is due to the fact that the proposed prior depends on the main direction of the observed intensity, i.e., $\mathbf{S}_{nu}$. This novel approach provides a pixel-wise-based estimation strategy and initializes a transmission involving many details corresponding to imaging contents. Moreover, if $\tilde{t}$ is used directly for computing the latent scene radiance $\mathbf{J}(x)$, the recovered result's contrast is low. Since the transmission map $\tilde{t}$ is obtained by the pixel-to-pixel computation, it will present the details of scenery which do not make sense. It is reasonable and necessary to make $\tilde{t}$ in (\ref{estimatedtransformula}) less structured and more smooth. Although there are many methods to smooth it, in this paper, we first downsample $\tilde{t}$ and then upsample it to obtain a smoothed $\tilde{t}$. This direct strategy can help get satisfactory results.

\subsection{Final recovery formula}\label{frf}
 Using the image formulation model (\ref{eq:IMF}), the final latent radiance $\mathbf{J}$ is given by
 \begin{equation}\label{comput_J}
    \mathbf{J}(x) = \frac{\mathbf{I}(x) - \omega\tilde{t}(x) \mathbf{A}}{\max(1-\omega \tilde{t}(x),t_0)},
 \end{equation}
where $\omega \in (0,1]$ is an introduced constant relaxation parameter and $t_0$ is the lower bound for a stable computation. We choose  $t_0 = 0.001$ in this paper. 

The estimation of the ambient light $\mathbf{A}$  is also very important. Note that the observed pixel is correlated with the ambient light, i.e., the pixel with the highest norm of the transmission is dominated by $\mathbf{A}$. To estimate $\mathbf{A}$,   we first pick the pixels with the highest top 0.1\% norm in $\tilde{t}$ to avoid outliers and then set the mean value of these pixels in the observed image $\textbf{I}$ to be  $\mathbf{A}$. The whole process of scene recovery by our method is shown in Figure \ref{fig:flowchart}.

\subsection{Complexity of our algorithm}
From the flowchart shown in Figure \ref{fig:flowchart}, we know that our method is quite straightforward, and we only need several simple steps. We need Eq. (\ref{Iu}) to compute the mean value of the channel images, and Eq. (\ref{estimatedtransformula}) to compute the transmission,  and a Gaussian filter to smooth on the image.
The last step is Eq. (\ref{comput_J}). All these steps are of complexity $O(N)$ where $N$ is the size of the single image. Hence, the complexity of our algorithm is  $O(N)$.

\section{Numerical Experiments}

In this section, we will elaborate on the implementation details, and analyze the imaging results on three different exemplary applications of the proposed method, i.e., sandstorm image enhancement, underwater image enhancement, and image dehazing.

 \subsection{Implementation}
The prominent advantage of our method is that the imaging performance only depends on $\omega \in (0, 1]$ in Eq. (\ref{comput_J}). The estimation of the transmission $\tilde{t}$ is thus parameter-free and robust under different visibility conditions. We have implemented numerous experiments to determine the optimal parameter $\omega$. This parameter is empirically set as $\omega = 0.8$, which guarantees high-quality visual results in most cases. Note that this parameter is application-based, the readers can try different $\omega$ to get a better performance.

All experiments are performed using Matlab R2018a on a machine with an Intel(R) Core (TM) i9-10850K CPU @3.60GHz. Unless specified, all test images are realistic scenarios collected from the Internet. Our method will be compared with several state-of-the-art methods. For the sake of fairness, the competing methods produce the most satisfactory imaging results with the best tuning parameters in this work.

 \subsection{Visibility restoration in sandstorm weather}
We first validate the effectiveness and robustness of our method for visibility reconstruction in sandstorm weather. The sandstorm could tremendously degrade the visibility due to the light scattering and absorption by floating sand and dust. It negatively influences practical applications, e.g., video surveillance system, automatic navigation, and remote sensing, etc. Performance of the proposed method is compared with six different approaches, i.e., DCP \cite{he2010single}, MSCNN \cite{ren2016single}, Haze-Lines \cite{berman2017air}, LDCP \cite{zhu2018haze}, Fusion \cite{fu2014fusion}, and Retinex \cite{fu2014retinex}. The sample sandstorm images and respective restoration results are visually displayed in Figure \ref{fig:sandimages}. It is observed that the comparative dehazing methods (i.e., DCP \cite{he2010single}, MSCNN \cite{ren2016single}, Haze-Lines \cite{berman2017air}, and LDCP \cite{zhu2018haze}) tend to restore the prominent structures, but fail to effectively suppress the impact of dust and sandstorms. This phenomenon could be related to the fact that images in sandstorm and haze weathers are generated based on different imaging theories. Although Retinex-based method \cite{fu2014retinex} achieves good performance, it still suffers from color distortion, i.e., cold-tone appearance. In contrast, our method can produce more natural-looking results with better structures in a more robust manner. Although all competing methods illustrate the capacity for sandstorm reduction, Retinex-based method \cite{fu2014retinex} has obvious color distortion problems, e.g., inconsistent brightness and loss of textural details, leading to image quality degradation. Due to the rank-one transmission prior, our method shows a stronger ability to robust visibility restoration in the sandstorm weather. 

To facilitate the related researches, we will release a dataset including realistic sandstorm images with different imaging scenes and sizes.

{\setlength{\parskip}{1em}
 \textbf{Run-time performance}. To test the efficiency of different methods on different single image sizes, we also show the run-time of our method on a machine with an NVIDIA GTX 2080Ti GPU (11GB RAM). 
 }

 Table \ref{tab:run_time} shows that our method performs much faster than competitors (even deep learning-based methods). With the GPU acceleration,  our method can reach real-time performance on the moderate size image.

Besides the astonishing performance on sandstorm image enhancement, our method also does well in underwater image enhancement and image dehazing. We will present some results in the following subsections.

\begin{table*}
    \caption{Run-time (seconds) performance. All methods are tested on the same machine using the same test images. Our method can reach real-time performance for the moderate image size using GPU (Nvidia 2080Ti). Without specified, all methods are carried out using CPU (lntel(R) Core(TM) i9-10850K CPU @3.60GHz). (Blue: Fastest method using CPU; Red: GPU acceleration on our method.)}  

    \label{tab:run_time}
    \vspace{0.2cm}
    \centering
    \scalebox{0.72}{
    \begin{tabular}{|c|c|c|c|c|c|c|c|c|c|c|c|}
    \hline
        Format  & DCP \cite{he2010single} & Retinex \cite{fu2014retinex} & LDCP \cite{zhu2018haze} & Fusion \cite{fu2014fusion} & MSCNN \cite{ren2016single} & Haze-Lines \cite{berman2017air}  & DehazeNet \cite{cai2016dehazenet}  & AODNet \cite{li2017aod} & CAP \cite{zhang2016dehazing} & Ours & Ours (GPU) \\ \hline
        360p & 0.47 &0.25 &  0.44& 0.28  & 0.55 & 0.39  & 0.85   & 0.36  & 0.54  & \textcolor{blue}{0.05}  & \textcolor{red}{0.04}  \\ \hline
        480p & 0.95 & 0.46& 0.86 & 0.54  & 0.98  & 0.83  & 1.88  &  0.70  &   0.78 & \textcolor{blue}{ 0.12}  & \textcolor{red}{0.04}  \\ \hline
        720p  & 2.51  & 1.20  & 2.29  & 1.57  & 2.48  & 2.47  & 6.01 & 1.84 & 1.56 & \textcolor{blue}{0.33}  & \textcolor{red}{0.07}  \\ \hline
        1080p  & 5.71 & 3.15  & 5.11  & 3.70  & 5.62  & 5.98   & 14.89 & 4.34 & 3.11 & \textcolor{blue}{0.80}  & \textcolor{red}{0.12}  \\ \hline
        2k  &  10.15  & 6.08  & 9.34  & 6.75  & 11.22  & 11.20   & 27.33 & 8.07 & 6.80 & \textcolor{blue}{1.47}  & \textcolor{red}{0.18}  \\ \hline
        4k & 27.75  & 17.58  & 25.42  & 18.60  & 30.90 & 35.54  & 76.22 & 21.72 & 13.55 & \textcolor{blue}{4.05}  & \textcolor{red}{0.46}  \\ \hline
    \end{tabular}
    }
\end{table*}

\subsection{Underwater image enhancement}

The realistic underwater images are obtained from a dataset named Underwater Image Enhancement Benchmark (UIEB) established by Li et al. \cite{li2019underwater}. We compare our method with two-step based method \cite{fu2017two}, blurriness-based method \cite{peng2017underwater}, UWCNN \cite{li2020underwater} and  one commercial application for enhancing underwater images (Dive+\footnote{https://itunes.apple.com/us/app/dive-video-color-correction/id12515 06403}). For the Dive+, we adjust the parameter settings to generate satisfactory results. Figure \ref{fig:underwatercmp} (a) is the raw underwater image that seems greenish. Without a doubt, such color deviation affects the visual quality. As can be seen from Figures \ref{fig:underwatercmp} (b)$\sim$(h), our method produces quite competitive results. Among these competitors, the two-step based method and the commercial app Dive+ perform better than others. Our results have better colorfulness, sharpness, and contrast.

\subsection{Image dehazing}
 We compare our method with the state-of-the-art that include DCP \cite{he2010single}, MSCNN \cite{ren2016single}, LDCP \cite{zhu2018haze}, and AODNet \cite{li2017aod}. The test images are obtained from the dataset (RESIDE) established by Li et al. \cite{li2018benchmarking}.  The DCP is a well-known prior-based method. However, DCP may fail when there exist large sky regions.  The luminance model (LDCP) can help assist in estimating the transmission map in the sky. AOD-Net \cite{li2017aod} directly restored the latent sharp image from a hazy image through a light-weight CNN.  A multi-scale CNN (MSCNN) is the first method that generates the transmission matrix by a coarse-to-fine strategy. Figure \ref{fig:dehazecmp} shows the comparisons of different methods. 
 
 In Figure \ref{fig:dehazecmp}, we find that DCP produces garish results, i.e., the sky presents abnormally blue. Although LDCP performs better than DCP, the results appear unreal. Both deep learning-based methods MSCNN and AODNet produce good and stable results. Visually, the dehazed results by our method are very competitive in comparison with others. The sky in our results looks more natural and real.

\section{Analysis and Discussion}\label{ana_dis}
The proposed rank-one prior provides a good guidance for computing the transmission by projecting the observed spectrum onto the unified spectrum.  Our method derives the transmission from the ambient light instead of the scene radiance. Using statistical validation on real haze datasets, we introduce rank-one prior and unified spectrum to describe and measure the ambient-light-based transmission. Hence, we approximate the transmission linearly by the unified spectrum for each pixel.

\subsection{Relation with other methods}
Both DCP and our method are based on physical assumptions. However, DCP is prior for the natural image, while ours is for the ambient light. Apparently, the transmission computed by DCP satisfies our rank-one prior.

In addition, our method is related to some automatic white balancing algorithm. Indeed, estimating $\mathbf{S}_u$ is similar to the gray world (GW) \cite{lukac2009single}, which requires illumination estimation. In GW, the mean of the r, g, b channels in a given scene should be roughly equal. It computes the average of individual red, green, and blue color components.

\subsection{Limitations}
Since we use the whole degraded image to obtain the unified spectrum, it may not be good enough if the distance between nearby object and distant object is too large. A possible solution is to use a supervised strategy, i.e., select a good region to obtain the unified spectrum. Another problem is that the originally unnoticeable impurities, such as noise and blocking, will be amplified \cite{chen2016robust,yan2013dense}.

\section{Conclusion}
In this paper, we have proposed a new and straightforward but effective method for the scene recovery in different real applications. The main idea of our work was that we considered an intensity projection strategy to estimate the transmission. This strategy was motivated by a rank-one transmission prior. The complexity of transmission estimation is $O(N)$ where $N$ is the size of the single image. The transmission map of a single degraded image obtained by the proposed method is very competitive with the state-of-the-art. Indeed, we can recover the scene in real-time. Supervised learning-based methods have achieved astonishing performance in the low-level computer vision field. The limitation of these methods is that they require a large amount of training data to improve the network's generalization ability and robustness. Although some self-supervised learning-based methods no longer rely on a large number of paired datasets to obtain results similar to supervised learning-based methods, limited by the sandstorms imaging models and related datasets, the research and development of deep learning-based sandstorm removal methods are progressing slowly. In contrast, our method has the best performance to a great extent. More results can be found in the supplemental material.

\section{Acknowledgements}
The work of Jun Liu  was supported by NSFC (11701079, 12071069, 11690012, 11631003) and Fundamental Research Funds for the Central Universities (2412020FZ023); The work of Ryan Wen Liu was supported by NSFC (51609195); The work of Jianing Sun was supported by Fundamental Research Funds for the Central Universities (2412019FZ032) and National Key R\&D Program (2020YFA0714100);  The work of Tieyong Zeng was supported by the CUHK startup, and the CUHK DAG under grant (4053342, 4053405), RGC (14300219,14302920), and grant (NSFC/RGC N$\_$CUHK 415/19). The authors would like to thank the three anonymous reviewers for their professional comments and constructive suggestions. We deeply thank Yuxu Lu in Wuhan University of Technology who kindly performs most of experiments in this work.

{\small
\bibliographystyle{ieee_fullname}

%\bibliography{egbib}
}

\end{document}